\documentclass{article}
\usepackage{graphicx}

\usepackage{nac_preprint}
\usepackage[utf8]{inputenc} 
\usepackage[T1]{fontenc}    
\usepackage{hyperref}       
\usepackage{url}            
\usepackage{booktabs}       
\usepackage{amsfonts}       
\usepackage{nicefrac}       
\usepackage{microtype}      

\usepackage{booktabs} 
\usepackage{subcaption}
\usepackage{amsmath}
\usepackage{amssymb}
\usepackage{cleveref}
\usepackage{mathtools}
\usepackage[toc,page]{appendix}
\usepackage{enumitem}
\usepackage{graphics}
\usepackage{graphicx}
\usepackage{xcolor, colortbl}
\usepackage[export]{adjustbox}
\usepackage{algorithm}
\usepackage{algorithmicx}
\usepackage{multicol}
\usepackage{algpseudocode}
\usepackage[inkscapelatex=false]{svg} 

\usepackage{longtable} 

\definecolor{LightCyan}{rgb}{0.88,1,1}
\definecolor{LightOrange}{rgb}{1,0.76,0.86}

\algnewcommand{\LineComment}[1]{\State \(//\) #1}
\algnewcommand{\RLineComment}[1]{\State \(\triangleright\) #1}
\newlength{\algrhswidth}
\usepackage{bm}
\usepackage{multirow}
 
\usepackage{wrapfig}

\usepackage{etoolbox}
\usepackage{tikz}
\usetikzlibrary{tikzmark}
\usetikzlibrary{calc}

\errorcontextlines\maxdimen

\newcommand{\ALGtikzmarkcolor}{black}
\newcommand{\ALGtikzmarkextraindent}{4pt}
\newcommand{\ALGtikzmarkverticaloffsetstart}{-.5ex}
\newcommand{\ALGtikzmarkverticaloffsetend}{-.5ex}
\makeatletter
\newcounter{ALG@tikzmark@tempcnta} 

\newcommand\ALG@tikzmark@start{%
    \global\let\ALG@tikzmark@last\ALG@tikzmark@starttext%
    \expandafter\edef\csname ALG@tikzmark@\theALG@nested\endcsname{\theALG@tikzmark@tempcnta}%
    \tikzmark{ALG@tikzmark@start@\csname ALG@tikzmark@\theALG@nested\endcsname}%
    \addtocounter{ALG@tikzmark@tempcnta}{1}%
}

\def\ALG@tikzmark@starttext{start}
\newcommand\ALG@tikzmark@end{%
    \ifx\ALG@tikzmark@last\ALG@tikzmark@starttext
    \else
        \tikzmark{ALG@tikzmark@end@\csname ALG@tikzmark@\theALG@nested\endcsname}%
        \tikz[overlay,remember picture] \draw[\ALGtikzmarkcolor] let \p{S}=($(pic cs:ALG@tikzmark@start@\csname ALG@tikzmark@\theALG@nested\endcsname)+(\ALGtikzmarkextraindent,\ALGtikzmarkverticaloffsetstart)$), \p{E}=($(pic cs:ALG@tikzmark@end@\csname ALG@tikzmark@\theALG@nested\endcsname)+(\ALGtikzmarkextraindent,\ALGtikzmarkverticaloffsetend)$) in (\x{S},\y{S})--(\x{S},\y{E});%
    \fi
    \gdef\ALG@tikzmark@last{end}%
}

\apptocmd{\ALG@beginblock}{\ALG@tikzmark@start}{}{\errmessage{failed to patch}}
\pretocmd{\ALG@endblock}{\ALG@tikzmark@end}{}{\errmessage{failed to patch}}
\makeatother

\usepackage{pifont}
\newcommand{\cmark}{\textcolor{blue}{\ding{52}}}%
\newcommand{\xmark}{\textcolor{red}{\ding{55}}}%

\title{R-AIF: Solving Sparse-Reward Robotic Tasks from Pixels with Active Inference and World Models}

\author{%
Viet Dung Nguyen\\
 Rochester Institute of Technology \\ 
\texttt{vn1747@rit.edu}
\And 
Zhizhuo Yang \\
Rochester Institute of Technology \\
\texttt{zy8981@rit.edu}
\And 
Christopher L. Buckley \\
VERSES AI Research Lab\\
University of Sussex\\
\texttt{c.l.buckley@sussex.ac.uk}
\And 
Alexander Ororbia \\
Rochester Institute of Technology \\
\texttt{ago@cs.rit.edu}
}

\begin{document}

\setlength{\abovedisplayskip}{0.065cm}
\setlength{\belowdisplayskip}{0pt}

\maketitle

\begin{abstract}
Although research has produced promising results demonstrating the utility of active inference (AIF) in Markov decision processes (MDPs), there is relatively less work that builds AIF models in the context of environments and problems that take the form of partially observable Markov decision processes (POMDPs). In POMDP scenarios, the agent must infer the unobserved environmental state from raw sensory observations, e.g., pixels in an image. Additionally, less work exists in examining the most difficult form of POMDP-centered control: continuous action space POMDPs under sparse reward signals. In this work, we address issues facing the AIF modeling paradigm by introducing novel prior preference learning techniques and~\textit{self-revision} schedules to help the agent excel in sparse-reward, continuous action, goal-based robotic control POMDP environments. Empirically, we show that our agents offer improved performance over state-of-the-art models in terms of cumulative rewards, relative stability, and success rate. The code in support of this work can be found at \url{https://github.com/NACLab/robust-active-inference}. 

\end{abstract}

\section{Introduction}
\label{sec:intro}

Reinforcement learning (RL) has notably been widely utilized in robotic systems to solve a variety of manipulation and control tasks~\cite{Russell2010AIModernBook, Han2023RLRobotSurvey, Morales2021RLRobotSurvey} using model-free RL algorithms such as the soft actor critic~\cite{Haarnoja2018SAC, Haarnoja2018SACAPP} or the deep Q-network~\cite{Mnih2013DQN}. Model-based RL, on the other hand, predicts the dynamics of a Markov decision process (MDP) and utilizes this learned generative model to plan a useful policy. From the `Dyna' framework~\cite{Sutton1991Dyna}, model-based RL has evolved into powerful modern-day models, including latent dynamics models~\cite{Bishop2006MLBook}, imagination-augmented RL~\cite{Racaniere2017ModelBasedRLAugmented}, or recurrent-based state dynamics models capable of playing Atari games~\cite{Kaiser2020ModelBasedRLAtari}. Within this domain of research, there exists a sub-field -- formally known as active inference -- that develops models that align with underlying principles of model-based RL. Generally, an active inference agent maintains and adapts a best estimate of its world (generative world model). It further aims to take actions that lead to outcomes that are aligned with its `preferences' while working to predict and minimize the degree of surprise it would potentially encounter in future engagements within its niche~\cite{Parr2022AIFFEPBook, Friston2003InferenceBrain}. However, current active inference research only tackles MDP robotics problems, with far less consideration for pixel-based POMDP tasks that elicit sparse reward signals; see Table~\ref{tab:features}.

\begin{table}[!t]
    \centering
    \begin{tabular}{l||c|c|c|c|c}
        Agent Model
    & \begin{tabular}[t]{@{}l@{}}
        POMDP
    \end{tabular}
     &\begin{tabular}[t]{@{}l@{}}
        Sparse\\ 
        Reward
    \end{tabular} 
     &\begin{tabular}[t]{@{}l@{}}
        Var.\\ 
        Goal
    \end{tabular} 
     &\begin{tabular}[t]{@{}l@{}}
        Cont.\\ 
        Action
    \end{tabular} 
     &\begin{tabular}[t]{@{}l@{}}
        Discrete\\ 
        Action
    \end{tabular}   
     \\ \hline\hline
        \cite{tschantz2020reinforcement, Tschantz2020Scaling}& \xmark & \cmark & \xmark & \cmark & \xmark \\
        \cite{Noel2021AIFCapsule} & \xmark & \cmark & \xmark & \cmark & \xmark \\
        \cite{Hafner2020Dreamerv1, hafner2020dreamerv2, hafner2023dreamerv3} & \cmark & \xmark~$^a$ 
        & \cmark & \cmark & \cmark \\ 
        \cite{vanderHimst2020AIFPOMDPCartPole} & \xmark~$^b$ 
        & \xmark & \xmark & \xmark & \cmark \\ 
        \cite{fountas2020DAIFMC}& \cmark & \xmark & \xmark & \xmark & \cmark \\
        \cite{millidge2020VariationalPolicyGradients} & \xmark & \xmark & \xmark & \xmark & \cmark\\
        \cite{Hoeffelen2021DAICarRacing}& \cmark & \xmark & \xmark & \xmark & \cmark \\
        \cite{Ccatal2019BayesianPolicyAIF}, \cite{Ccatal2020AIFPOMDP}& \cmark~$^c$
        & \cmark & \xmark & \xmark & \cmark \\
        \cite{Mazzaglia2021ContrastiveAIF}& \cmark & \cmark~$^d$
        & \xmark~$^d$
        & \cmark & \cmark \\ \hline 
        \textbf{R-AIF (ours)} & \cmark & \cmark & \cmark & \cmark & \cmark
    \end{tabular}
    \caption{Features (``var.'' for ``variable'' and ``cont.'' for ``continuous'') of different model-based reinforcement learning and deep active inference research efforts. \\
    \footnotesize{
    $^a$ Dreamer is not specifically designed to work with sparse-reward problems, but was found in this work to be robust enough to score decently well. \\ 
    $^b$ This model was designed specifically for the Cartpole environment with a modified observation space. Furthermore, the image observation is stacked for four steps, making it a MDP-like problem~\cite{mnih2015human}.\\
    $^c$ This work forms a non-pixel-level POMDP by hiding the velocity state, only presenting the position of the mountain car as the observation (and does not work with raw sensory data).\\ 
    $^d$ Although this agent is designed for sparse rewards, providing the goal image is not practical for robot environments with varied goal states (as we study in this work).
    }
    }\label{tab:features}
\end{table}

In this work, we make the following key contributions. 
\textbf{1)} We propose a novel \textit{contrastive recurrent state prior preference} (CRSPP) model, which allows the agent to learn its own preference over the world's state(s) online. This online preference dynamically shapes the agent's policy distribution, improving general performance.
\textbf{2)} We propose a new formulation of expected free energy and optimize it using the actor-critic method, improving the stability of the action planner compared to other active inference baselines. 
\textbf{3)} We propose our \textit{robust active inference} (R-AIF) agent with a~\textit{self-revision mechanism}, demonstrating its potential to improve over general RL methods, boosting the model convergence rate in sparse-reward tasks. Finally, 
\textbf{4)} We provide empirical evidence that our R-AIF agent converges faster and is more stable than a state-of-the-art model-based reinforcement learning baseline (DreamerV3~\cite{hafner2023dreamerv3}) and powerful active inference baselines: DAIVPG~\cite{millidge2020VariationalPolicyGradients} and the agent in~\cite{vanderHimst2020AIFPOMDPCartPole}.

\section{Related Work}
\label{sec:lit_review}

\textbf{Model-based reinforcement learning} (MBRL) is a pivotal approach in reinforcement learning that centers around the `world model', a concept that involves creating an internal model of the environment to guide the agent's future actions. This approach is exemplified in works related to state space models~\cite{Buesing2018RLGenerativeModel}, `embed to control'~\cite{Watter2015E2C}, `Plan2Explore'~\cite{SekarPlan2Explore}, `dreaming'~\cite{Okada2021Dreaming}, `PlaNet'~\cite{Hafner2019Planet}, `Dreamer'~\cite{Hafner2020Dreamerv1}, divergence minimization~\cite{Hafner2020ActionAP}, and perceptual uncertainty~\cite{pathak2017curiosity, Shyam2018MAX}.

\textbf{Active inference} (AIF) is a framework in cognitive science and neuroscience which centers around the notion of generative models which ``understand'' a ``lived world''~\cite{Parr2022AIFFEPBook, Pezzulo2024InteractWithWorld} or (eco)niche. Importantly, AIF itself is in effect a corollary of the free energy principle, which posits that biological systems minimize a quantity known as free energy as they continuously work to preserve their existence (self-evidence) and interact with their environments effectively. Formally, AIF involves minimizing the quantity known as variational free energy which is formally defined as follows:
\begin{align}
    F = \mathbb{E}_{Q(s)}[\ln Q(s) - \ln P(o, s)]
\end{align}
where $Q(s)$ is the approximate posterior over states $s$ and $P(o, s)$ is the joint probability of states $s$ and observations $o$. This framework extends to machine intelligence since it casts the perception and planning problem as a Bayesian inference problem where an agent updates its beliefs about the sensory inputs and selects actions that lead to the minimization of expected free energy (EFE):
\begin{align}
    G(\pi) = \mathbb{E}_{Q(s)} [ \ln Q(s, \theta | \pi) - \ln Q(o, s, \theta | \pi ) ]
\end{align}
where $\theta$ contains the model parameters, $\pi$ is the planned action distribution, $Q(s, \theta | \pi)$ is the approximate posterior over state and model parameters given the agent's actions, and $Q(o, s, \theta | \pi )$ is the approximate posterior over observation, state, and model parameters given the agent's actions. The EFE is often broken down in two key terms: instrumental and epistemic. The instrumental term defines how future estimated states/observations align with the agent's prior preference (interest) given its current action plan, whereas the epistemic term describes the surprise/uncertainty level associated with the estimated future states under a given planned policy. Generally, this AIF framing offers a principled basis for RL models that learn and act by reducing the mismatch between predicted and observed data. Deep active inference seeks to utilize the tools of deep learning, e.g,. deep neural networks and backpropagation of errors, to make estimation/production of the core values inherent to EFE easier to simulate efficiently; see \cite{Mazzaglia2022DeepAIFSurvey} for a review.

\section{Solving Sparse-Reward Pixel Robotic Tasks}\label{sec:method}

In order to tackle a robotic task with a continuous action space, varied goals, and sparse reward signals, we first re-formulate the construction of the AIF/MBRL agent's generative world model~\cite{Ha2018WorldModel, friston2017AIFProcessTheory} in Section~\ref{sec:method-wm} (while Section~\ref{sec:method-crspp} introduces our novel contrastive recurrent state prior preference model). We propose the~\textit{robust active inference}\footnote{Implementation details available at \url{https://github.com/NACLab/robust-active-inference}} (R-AIF) agent in Section~\ref{sec:method-raif}, which utilizes actor-critic methodology~\cite{Sutton2018RLBook} in order to optimize the action `planner', ultimately seeking to minimize both instrumental and epistemic signals from its learned world model.

Our agent operates on standard POMDPs, in discrete time, where the interaction between the agent and the environment is formally expressed as $\mathcal{M} = \left(\mathcal{S}, \mathcal{A}, \mathcal{O}, p, E, r, g \right)$. $\mathcal{S}$ is the set of all environment states (hidden from the agent), $\mathcal{O}$ is the set of observations, $E(o_t|s_t)$ is an emission function which produces the observable signal $o_t$ conditioned on the unobserved state distribution $s_t$. Additionally, we have the action space $\mathcal{A}$, the reward function $r: \mathcal{S} \times \mathcal{A} \rightarrow \mathbb{R}$, and the transition probability $p(s_{t+1}|s_t, a_t)$~\cite{Sutton2018RLBook}. Finally, we also consider the function $g: \mathcal{S} \rightarrow \{0,1\}$ indicating whether the agent has achieved its goal (or not) in a particular time step.

\begin{figure}[t]
    \centering
    \includegraphics[width=0.6\linewidth]{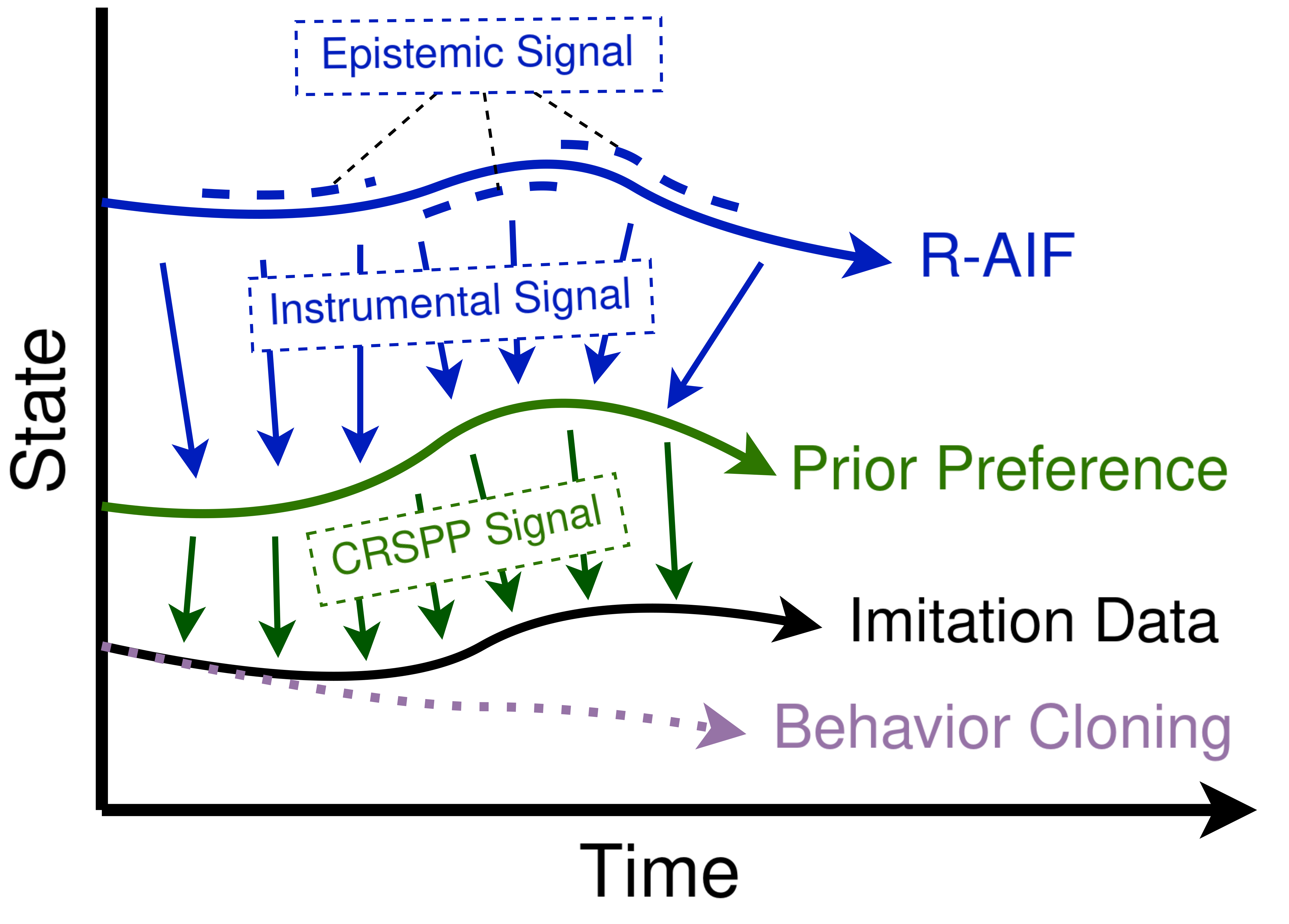}
    \caption{\textbf{Demonstration of different abstract trajectories with state (y-axis) through time (x-axis).} Behavior cloning trajectory diverges from the training trajectory due to small mistakes made by the agent as well as environmental stochasticity (due to the \textit{i.i.d} assumption applied to the environment). We instead want to estimate a ``preferred'' trajectory that closely matches the underlying data distribution. As a result, our R-AIF agent can ``nudge'' its trajectory toward its own prior preference. The dashed lines around the line R-AIF represents the epistemic signal of the agent that facilitates intelligent exploration within a safe range.}
    \label{fig:crspp-intuition}
\end{figure}

\subsection{The Generative World Model}\label{sec:method-wm}

Active inference posits that an agent finds an action sequence based on an estimated future state distribution~\cite{Friston2010FEP, Smith2022TutorialPaper, Friston2015Epistemic, yang2023neural, Friston2013LifeAsWeKnowIt, Millidge2021WhenceEFE}. To achieve this, we predict the next state $s_{t+1}$ given the current state $s_t$ and action $a_t$, along with a recurrent state $h_t$ that serves as temporal memory ($z_t$ serving as the memory's output). For modeling $h_t$, we employ a gated recurrent unit~\cite{chung2014gru} following the approach from~\cite{Noel2021AIFCapsule, Hafner2020Dreamerv1, hafner2020dreamerv2, hafner2023dreamerv3}. In general, this framing of the temporal integration of information is referred to as the \emph{recurrent state space model} (RSSM)~\cite{Hafner2019Planet}, which builds on concepts from the state space model literature~\cite{Buesing2018RLGenerativeModel, Ha2018WorldModel, Karl2017VariationalBayesFilters, Doerr2018PRSSM}. The RSSM formulation can formally be expressed as follows:
\begin{equation}\label{eqn:wm}
\begin{aligned}
    \text{Latent dynamics model: } &f_{\theta^\mathcal{R}}(z_t, h_t | s_{t-1}, a_{t-1}, h_{t-1}) \\
    \text{Approx. posterior (encoder): } &q_{\theta^\mathcal{E}}(s_t | o_t, h_t)\\
    \text{Prior (transition): } &p_{\theta^\mathcal{T}}(s_t|z_t) \\
    \text{Likelihood (decoder): } &q_{\theta^\mathcal{D}}(o_t | s_t, h_t)\\
    \text{Reward (decoder): } &q_{\theta^r}(r_t | s_t, h_t).
\end{aligned}
\end{equation}
Throughout this work, unless stated otherwise, every posterior $q(\cdot)$ and prior $p(\cdot)$ estimator that is parameterized by an artificial neural network (ANN) will be equipped with a recurrent neural network (RNN), i.e., a latent dynamics model. For clarity and simplicity, we have omitted the RNN and the module letter (i.e., $\mathcal{E}$ for encoder) in the notation; this leaves us with the encoder $q_{\theta}(s_t | o_t)$, transition $p_{\theta}(s_t|s_{t-1}, a_{t-1})$, decoder $q_{\theta}(o_t | s_t)$, and the reward predictor $q_{\theta}(r_t | s_t)$.

As the agent seeks to minimize its surprise, it continuously changes its belief(s) over hidden states to match its prior while maximizing the likelihood of observation~\cite{Smith2022TutorialPaper}; this is done by maximizing the evidence lower bound (ELBO)~\cite{Kingma2014VAE} employed in variational inference. The posterior over hidden state $q(s_t)$ can then be optimized by treating the agent's (free) energy function in terms of a gradient descent objective~\cite{Lecun2015DeepLearningBook, Goodfellow2016DeepLearningBook}. Minimizing the free energy based on both observed data $o_t$ and the prior transitioned state $p(s_t|s_{t-1}, a_{t-1})$ is often known as marginal free energy minimization~\cite{Parr2019MarginalMessagePassing}, closely related to the mechanisms of variational autoencoders~\cite{Kingma2014VAE} and stochastic variational inference procedures~\cite{Hoffman2013StochasticVI}:
\begin{equation}
\label{eqn:train-wm}
\begin{aligned}
    \operatorname*{arg\,min}_\theta \mathcal{L}_t(\theta) = &\underbrace{\text{D}_{\text{KL}}\left[ q_{\theta}(s_t | o_t) \parallel p_{\theta}(s_t | s_{t-1}, a_{t-1}) \right]}_{\text{complexity}} \\
    + &\underbrace{\mathbb{E}_{q_{\theta}(s_t|o_t)} \left[ -\ln \left( q_{\theta}(o_t | s_t) \right) \right]}_{\text{accuracy}}.
\end{aligned}
\end{equation}
Minimizing the complexity (term) aids the agent in closing the gap between its prior and its approximate posterior,  whereas minimizing the accuracy (term) improves the model's future observation estimation. We utilize the world model, which has a discretized state space~\cite{hafner2020dreamerv2} where each hidden state is represented by a vector of discrete distributions instead of a vector of Gaussian distributional parameters (as is done in other deep active inference formulations). We also employ the KL balancing trick~\cite{hafner2020dreamerv2} and apply the ``symlog'' function to inputs~\cite{hafner2023dreamerv3} for numerical stability.

\subsection{Contrastive Recurrent State Prior Preference (CRSPP)} 
\label{sec:method-crspp}

In active inference, the agent takes the actions that it believes would lead to its preferred outcomes (i.e. using the instrumental signal)~\cite{Sajid2021AIFDemystified, Millidge2021WhenceEFE, friston2017curiosity, Friston2010FEP, Friston2015Epistemic, friston2017AIFProcessTheory}. To construct this prior preference, past work has provided a goal state/observation(s) directly to the agent~\cite{Mazzaglia2021ContrastiveAIF} or manually crafted a prior preference distribution~\cite{friston2017AIFProcessTheory, yang2023neural}. However, the first approach suffers from sparsity over the preference space while the second is impractical in more realistic POMDPs. To tackle this, we leverage a small quantity of seed imitation data to learn an ANN that dynamically produces the preference over states at each time step; this effectively provides an easily-generated dense instrumental/goal signal. Concretely, we design the agent such that it moves according to a trajectory that is shaped towards its own estimation of future preferred states, ``nudging'' its own trajectory toward the imitation/positive data distribution (see Figure~\ref{fig:crspp-intuition}).

\begin{figure}[t]
    \centering
    \includegraphics[width=0.8\linewidth]{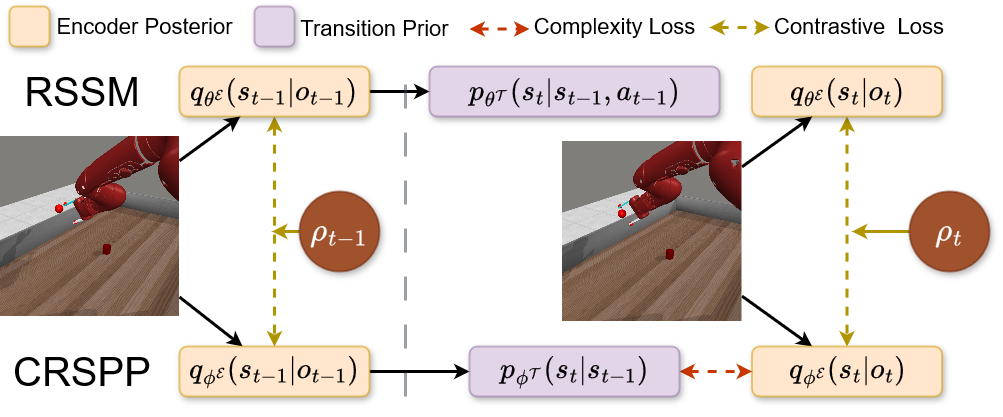}
    \caption{\textbf{The CRSPP learning framework.} CRSPP learns by optimizing the KL divergence between its approximate posterior and prior only when a state is ``desired'', i.e. $\rho_t > 0$. It also learns to predict next preferred states using a dynamic contrastive loss based on $\rho_t$ (which focuses on narrowing the gap between the estimated preferred state distribution and the actual approximate posterior produced by the RSSM).}
    \label{fig:crspp}
\end{figure}

\textbf{Dynamic Prior Preference Model Formulation.} In this work, we consider a prior preference model that takes in the image observation $o_t \in O$ and produces a posterior estimate over state $s_t \in S$. Based on this latent representation, the model then estimates -- or ``imagines'' -- the future latent representation that has a high preference value $\Tilde{s}_{\tau:H}$ over a time horizon $H$. To achieve this, we construct an RSSM without action encoded into the transition prior. Our model can then be further parameterized with an encoder posterior $q_{\phi}(s_t | o_t)$ and a transition prior $p_{\phi}(s_t | s_{t-1})$ (see Figure~\ref{fig:crspp}).

\textbf{Goal-Oriented Credit Assignment and Prior \textit{Self-Revision}.} Assuming that the trajectories collected throughout the R-AIF agent's learning process form a set $\mathbb{E} = \{e_0, e_1, \dots\}$, if we follow the conventions of contrastive learning methodology~\cite{laskin2020curl, he2020momentum, Oord2018CPC}, we partition this set of experiences into two portions based on the success status of each experience, i.e. $P(\mathbb{E})= (\{e_{+}\}, \{e_{-}\})$ where $\{e_{-}\} = \mathbb{E} \setminus \{e_{+}\}$. Intuitively, we aim to learn a prior preference model which estimates the preferred state $\Tilde{s}$ that is closer to (positive) states $s \in e_+$ while pushing $\Tilde{s}$ away from $s \in e_-$ (negative states). Specifically, to learn a model that performs roll-outs over a finite horizon with only preferred states, one can maximize the similarity of states between the prior preference model and the actual generative world model -- where ``reached goal states'' are of ``strong interest'' (yielding a positive signal) -- while minimizing this similarity measurement for the situations that the agent fails the task within an episode (yielding a negative signal). We consider a prior preference rate $\rho_t$, at every time step, that is positively-signed in successful episodes, and negatively-signed otherwise. This signal is decayed backward from the end of the trajectory in order to reward/penalize the immediate actions that led to success/failure within the task. Note that we set $\rho_t$ at its highest value at each successful state and decay this backward. In contrast, when the episode fails, the agent only needs to decay negatively backward from the end of the episode. We call this computation of $\rho$ the~\textit{self-revision mechanism}~\footnote{See footnote 1 for details.} and use this rate as a scalar for the contrastive objective -- facilitating a form of \textit{dynamic contrastive learning} -- when optimizing the CRSPP. As a result, our positive/negative scoring mechanism puts more weight on the states that are near the goal state, which partly resembles hindsight experience replay~\cite{Andrychowicz2017HER}.

\begin{figure}[t]
    \centering
    \begin{subfigure}{0.94\linewidth}
		\includegraphics[width=\linewidth]{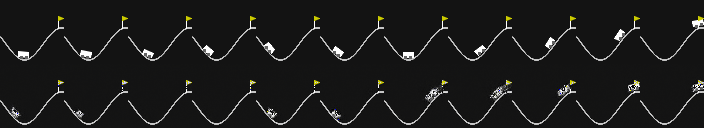}
		\label{fig:mtc_pref_traj}
	\end{subfigure}
	\begin{subfigure}{0.94\linewidth}
		\includegraphics[width=\linewidth]{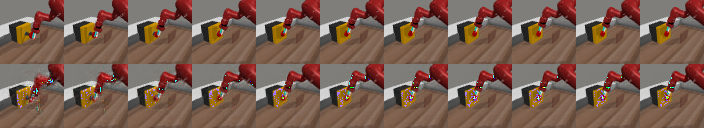}
		\label{fig:mtw_pref_traj}
	\end{subfigure}
    \caption{Actual observation (top row) versus the prior preference estimation (bottom row) across time (horizontal axis) of the mountain car problem (top image group) and the Meta-World `button press wall' task (bottom image group). We see that CRSPP produces a goal dynamically at each time step.}
    \label{fig:recon-preference}
\end{figure}

Note that we may learn the prior preference model using any contrastive objective that is conditioned on this prior preference rate. Additionally, we minimize the KL divergence between the prior and the posterior such that the prior preference model possesses an accurate positive (sample) imagination. In our work, we use cosine similarity, e.g., $\text{sim}(A, B) = \frac{A \cdot B}{\parallel A \parallel \parallel B \parallel}$, to optimize the CRSPP model:
\begin{equation}\label{eqn:train-crspp}
\begin{aligned}
    &\operatorname*{arg\,min}_\phi \mathcal{L}_t(\phi) = \text{ max} \bigl( 0, \text{sgn}(\rho_t) \bigr) \times \\
    & \text{ D}_{\text{KL}}\left[ q_{\phi}(s_t | o_t) \parallel p_{\phi}(s_t | s_{t-1}) \right] - \rho_t \text{ sim} ( \hat{s}_t, s_t )
\end{aligned}
\end{equation}
where $\hat{s}_t \sim q_{\phi}(s_t | o_t)$, $s_t \sim \text{sg}\bigl( q_{\theta}(s_t | o_t) \bigr)$, `sg' is the stop gradient function, `sgn' is the sign function, and $\rho_t$ is the prior preference rate computed at the end of each episode using the \textit{self-revision mechanism} (see Section~\ref{sec:method-crspp}). 
In general, the KL term helps in estimating the next preferred state prior more accurately, and the similarity term helps to dynamically push or pull the state space of the prior preference model in relation to the world model based on 
$\rho_t$. 
As a result, the prior preference model is trained to only produce the next set of preferred states/latent dynamics (without producing the failing states) based on $\rho_t$. 
Note that we further use the world model's decoder on the preferred state to produce the agent's goal image at each time step (see Figure~\ref{fig:recon-preference}). Being able to produce goals dynamically at each time step helps to shape the local prior preference towards an optimal trajectory, serving as a precursor to optimizing the action planner (while utilizing the gradient of the contrastive model).

\subsection{R-AIF Agent Behavioral Learning}
\label{sec:method-raif}

In active inference, the agent estimates both future states and observations and then plans action sequences based on the expected free energy computed from these future `imagined realities'~\cite{friston2017graphical, friston2016learning, friston2017AIFProcessTheory, fountas2020DAIFMC}. Although one can estimate both states and observations with respect to `imagination space', for practical model inference, it is also possible to roll-out only the latent dynamics into the future~\cite{Hafner2019Planet, Okada2021Dreaming, Rajeswar2023UnsupervisedRLPixels}; in this work, we roll out only latent states. Formally, with the planning time step $\tau$ and imagination horizon $H$, we estimate the future state $s_\tau \sim p_{\theta}(s_\tau | s_{\tau - 1}, a_{\tau - 1})$ using the (estimated) action $a_\tau \sim \pi_\psi (a_\tau | s_\tau)$. We train the agent to minimize the expected free energy based on these ``imagined'' future states. Similarly to~\cite{hafner2023dreamerv3}, we construct a policy network $\pi_\psi(a_\tau | s_\tau)$ that maximizes the estimated advantage from the value function $f_\chi (v_\tau | s_\tau)$ such that:
\begin{equation}
\begin{aligned}
    f_\chi (v_\tau | s_\tau) \approx \sum_{t=\tau}^{t+H}{ \gamma^{t - \tau} \left( r_\tau + \text{sim}(s_\tau, \hat{s}_\tau) - \text{IG}_\tau \right) }
\end{aligned}
\end{equation}
where the action $a_\tau$ is a $\tanh$-normalized sampled from the Gaussian distribution with a mean $\mu_\psi$ and standard deviation $\sigma_\psi$ produced by the policy network. The value function is then trained to estimate the reward $r_\tau$, similarity between the estimated future state $s_\tau$ and the imagined prior preference $\hat{s}_\tau$, and the negative information gain $-\text{IG}_\tau$.

\textbf{Expected Free Energy.} R-AIF's behavioral learning involves training the policy network to take actions that minimize the expected free energy~\cite{Schwartenbeck2019Curiosity, friston2017AIFProcessTheory}. We construct our formulation of the expected free energy as below:
\begin{subequations}
\begin{align}
    &G_\tau(\pi_\psi) = \ - \mathbf{H} [ \mathbb{E}_{Q(o_\tau, s_\tau)} Q(\pi_\psi | o_\tau, s_\tau)] \label{eqn:G_policy}\\
    & + \mathbf{H} [ \mathbb{E}_{Q(\theta , \pi_\psi)} Q(s_\tau | \theta, \pi_\psi)] - \mathbb{E}_{Q(\theta|\pi_\psi)} \mathbf{H} [ Q (s_\tau | \theta, \pi_\psi)] \label{eqn:G_ig}\\
    & - \mathbb{E}_{ Q(\theta | \pi_\psi) } \mathbf{H} [Q(s_\tau | \theta, \pi_\psi)] \label{eqn:G_state}\\
    & - r_\tau - \text{ sim} (s_\tau, \hat{s}_\tau).\label{eqn:G_inst}
\end{align}
\end{subequations}
Following AIF process theory, we aim to train our policy network $\pi_\psi$ to minimize the expected free energy $G_\tau(\pi)$, including the instrumental and epistemic signals. For the instrumental signal, the policy is trained to maximize the expected future estimated reward and the similarity between states and the agent's prior preference (Equation~\ref{eqn:G_inst}). For the epistemic signal, the policy network is provided with ``incentives'' when entering a state with higher policy entropy (Equation~\ref{eqn:G_policy}), similar to the processes used in the soft actor-critic frameworks~\cite{Haarnoja2018SAC}. Furthermore, the policy network is trained to reduce the model parameter's uncertainty (Equation~\ref{eqn:G_ig}) or information gain (IG)~\cite{Lindley1956IG, MacKay2003InformationTheoryBook}. This means that the agent takes more certain actions to maintain ``homeostasis''~\cite{Friston2006FEPBrain, ororbia2023mortal}. Similar to~\cite{tschantz2020reinforcement, Shyam2018MAX}, an ensemble of ANNs is utilized to computed IG (Equation~\ref{eqn:G_ig}). The final epistemic signal provides the agent with an intrinsic reward whenever the agent enters unknown states with a high entropy over future state predictions (in other words, generative world model entropy; Equation~\ref{eqn:G_state}). Maximizing this uncertainty about future states is equivalent to providing additional motivation~\cite{Barto2013IntrinsicMotivation, Oudeyer2007IntrinsicMotivation, Deci1985IntrinsicMotivationBook} for state exploration.

\begin{figure*}[h]
	\centering
	\begin{subfigure}{0.24\linewidth}
		\includegraphics[width=\linewidth]{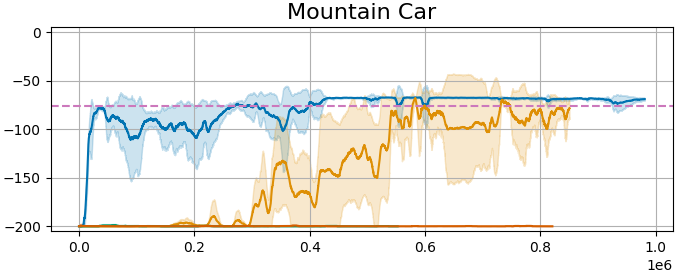}
	\end{subfigure}
	\begin{subfigure}{0.24\linewidth}
		\includegraphics[width=\linewidth]{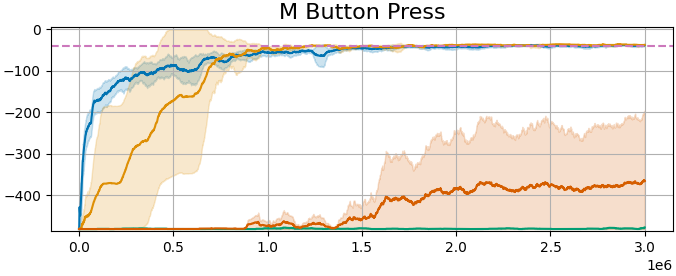}
	\end{subfigure}
	\begin{subfigure}{0.24\linewidth}
	        \includegraphics[width=\linewidth]{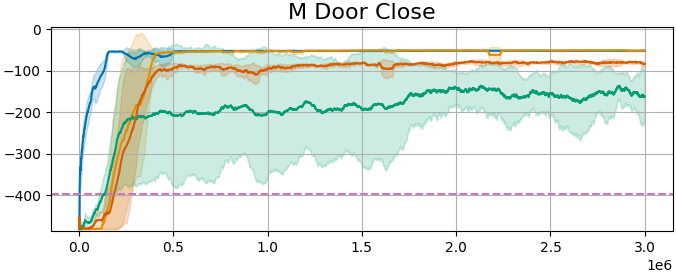}
         \end{subfigure}
    \begin{subfigure}{0.24\linewidth}
	        \includegraphics[width=\linewidth]{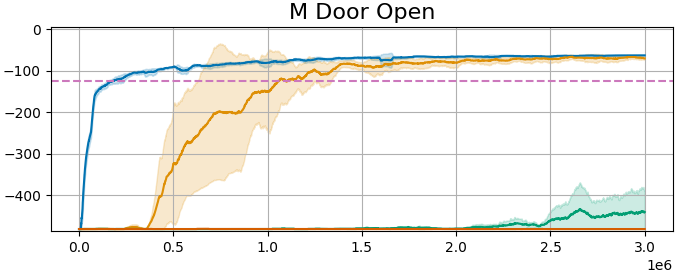}
         \end{subfigure}
 \begin{subfigure}{0.24\linewidth}
	        \includegraphics[width=\linewidth]{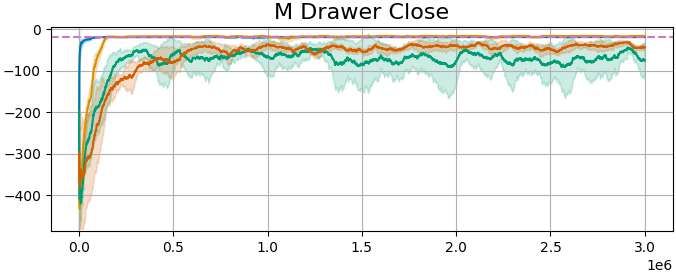}
         \end{subfigure}
 \begin{subfigure}{0.24\linewidth}
	        \includegraphics[width=\linewidth]{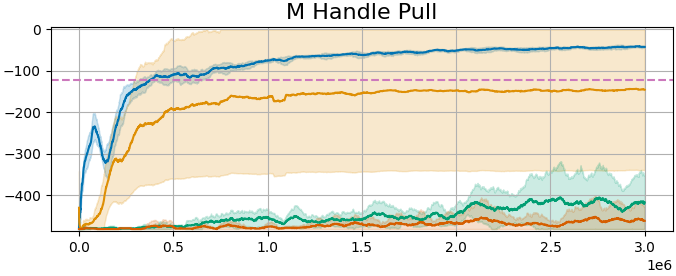}
         \end{subfigure}
 \begin{subfigure}{0.24\linewidth}
	        \includegraphics[width=\linewidth]{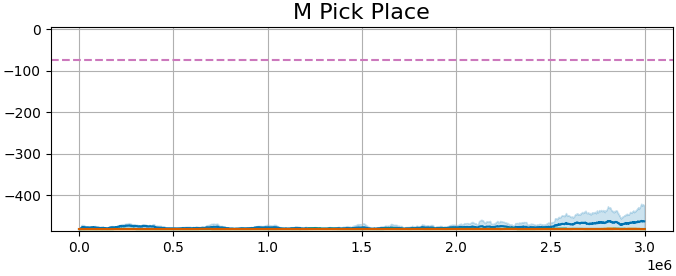}
        \end{subfigure}
 \begin{subfigure}{0.24\linewidth}
	        \includegraphics[width=\linewidth]{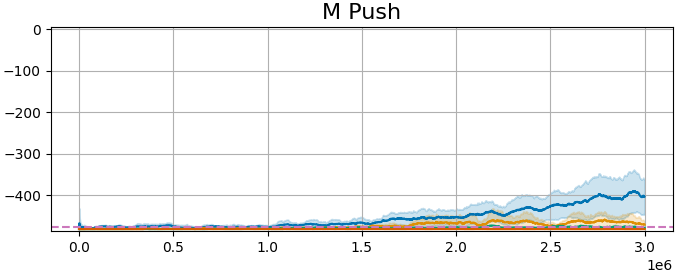}
         \end{subfigure}
 \begin{subfigure}{0.24\linewidth}
	        \includegraphics[width=\linewidth]{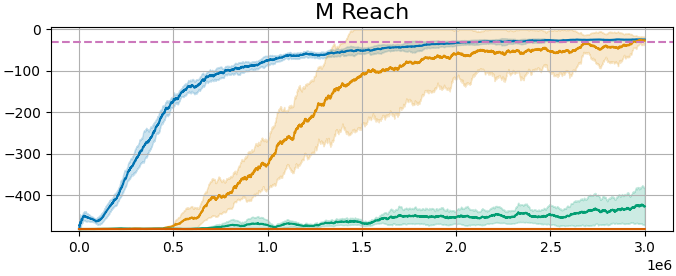}
         \end{subfigure}
 \begin{subfigure}{0.24\linewidth}
	        \includegraphics[width=\linewidth]{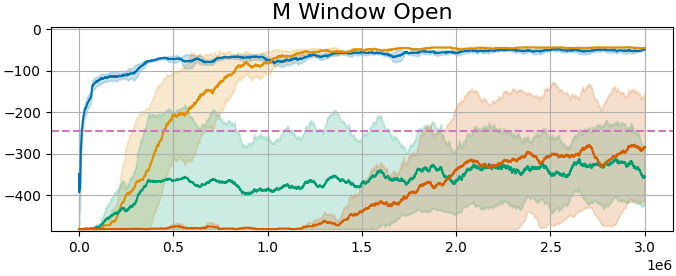}
         \end{subfigure}
         \begin{subfigure}{0.24\linewidth}
	        \includegraphics[width=\linewidth]{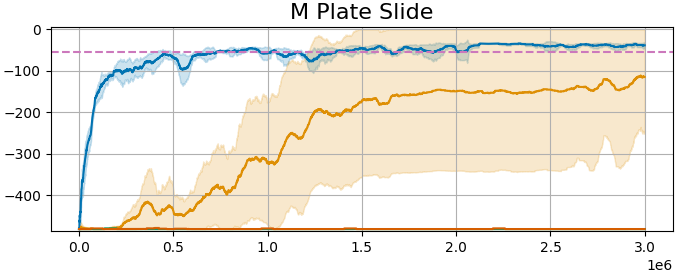}
         \end{subfigure}
         \begin{subfigure}{0.24\linewidth}
	        \includegraphics[width=\linewidth]{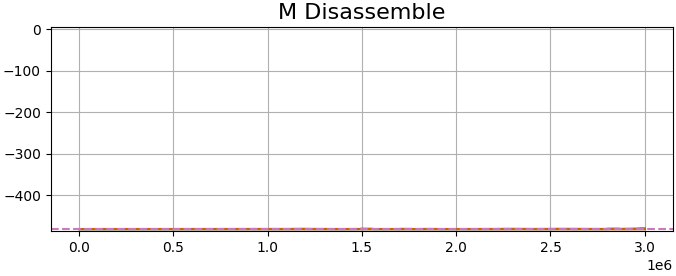}
         \end{subfigure}
         \begin{subfigure}{0.24\linewidth}
	        \includegraphics[width=\linewidth]{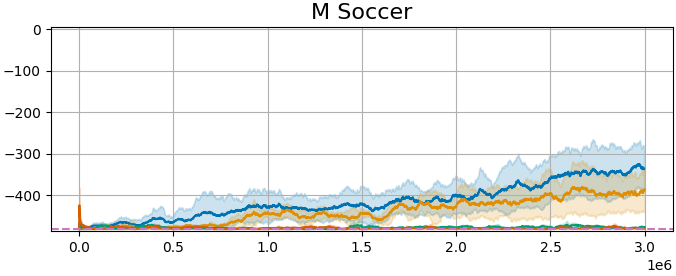}
         \end{subfigure}
         \begin{subfigure}{0.24\linewidth}
	        \includegraphics[width=\linewidth]{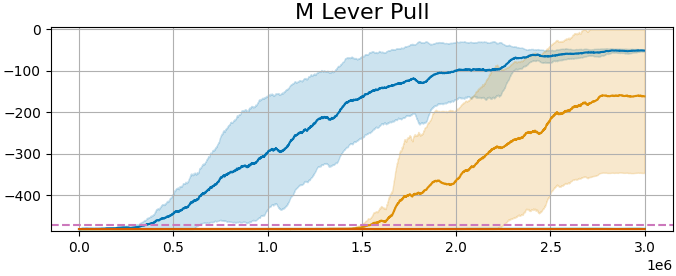}
         \end{subfigure}
         \begin{subfigure}{0.24\linewidth}
	        \includegraphics[width=\linewidth]{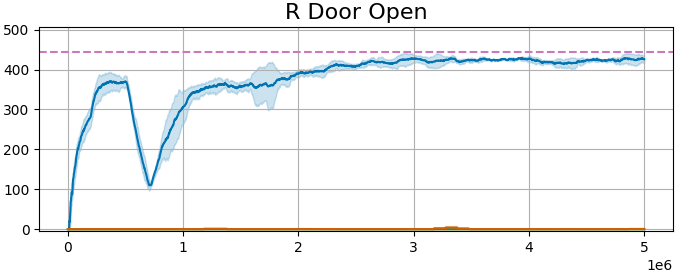}
         \end{subfigure}
         \begin{subfigure}{0.24\linewidth}
	        \includegraphics[width=\linewidth]{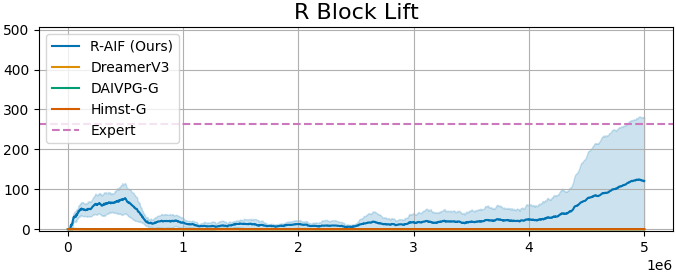}
         \end{subfigure}
	\caption{Cumulative reward ($y$-axis) trend through environment time steps ($x$-axis) of different agents. Pink dashed lines are average reward of the expert in the MDP version of the task.}
	\label{fig:reward-trends}
\end{figure*}

\textbf{Training Value Function.} We train the value network by minimizing the mean squared error (MSE) between the output of the value network and the target value computed at each time step. The target value is computed from the discounted cumulative rewards and information gain through generalized advantage estimate~\cite{Schulman2015GeneralGAELambda} and temporal difference~\cite{Sutton2018RLBook} methods, the use of which resembles sophisticated inference~\cite{Friston2021SophisticatedInference, Paul2023DPEFE}. Specifically, assuming that the agent is rolled-out over a state space with horizon $H$ and a future time step $\tau$, the target $\lambda$-return (value) for the advantage $\mathcal{G}$ is computed as follows:
\begin{equation}\label{eqn:G_value}
\begin{aligned}
    \mathcal{G}_\tau = &\left( r_\tau + \text{sim}\Bigl( s_\tau, \hat{s}_\tau \Bigr) - \text{IG}_\tau \right) \\
    + &\gamma c_\tau \Bigl( (1 - \lambda) f_{\chi}(v_\tau | s_\tau) + \lambda \mathcal{G}_{\tau+1} \Bigr)
\end{aligned}
\end{equation}
where $r_\tau \sim q_\theta( r_\tau | s_\tau), s_\tau \sim p_\theta(s_\tau | s_{\tau-1}, a_{\tau-1}), \hat{s}_\tau \sim \text{sg} \bigl( p_\phi(s_\tau | s_{\tau-1}) \bigr)$. $\mathcal{G}_H = v_H = f_{\chi}(v_\tau|s_\tau)_H$, and $\gamma$ is the discount factor. Similar to~\cite{hafner2023dreamerv3}, $c_\tau$ is an estimated boolean value that specifies whether or not an episode will continue. We can then optimize the $\lambda$-value estimator by minimizing the MSE between its estimation and the target as follows:
\begin{equation}\label{eqn:train-value}
\begin{aligned}
    \operatorname*{arg\,min}_\chi \mathcal{L}_\tau(\chi) = &\ \mathbb{E}_{p_\theta(s_\tau | s_{\tau-1}, a_{\tau-1}) \pi_\psi (a_{\tau-1} | s_{\tau-1})} \\
    &\left[ f_\chi (v_\tau | s_\tau) - \text{sg}\left(\mathcal{G}_\tau\right) \right]^2.
\end{aligned}
\end{equation}

\textbf{Behavioral Learning.} In this work, we train the policy network by minimizing its EFE $G_\tau(\pi_\psi)$: maximizing the policy entropy (Equation~\ref{eqn:G_policy}), the \textit{generative world model entropy} (Equation~\ref{eqn:G_state}), and the estimated advantage value as computed from the value function $\mathcal{G}_\tau \approx f_\chi(v_\tau | s_\tau)$. This advantage value is composed of an instrumental signal -- reward and similarity (\ref{eqn:G_inst}) -- and the negative information gain (Equation~\ref{eqn:G_ig}). Additionally, we also found the integration of an ``actor refresh'' term~\cite{Ororbia2023ActivePredictiveCoding}, e.g., $- \mathbb{E}_{\pi_\psi (a_\tau | s_\tau)} \ln (a_\tau^*)$, to be useful in ensuring good performance. In general, the agent learns to shift its trajectory toward the preferred state distribution, taking actions that it is confident in while exploring uncertain states (see Figure~\ref{fig:crspp-intuition}). Formally, we aim to maximize the actor's objective functional as follows:
\begin{equation}\label{eqn:train-actor}
\begin{aligned}
    &\operatorname*{arg\,max}_\psi \mathcal{L}_\tau(\psi) = \mathbb{E}_{\Tilde{Q}(a, s | \psi, \theta)} \Bigl[ f_\chi (v_\tau | s_\tau) \\
    &+ \zeta \mathbf{H}\left[ p_\theta(s_\tau | s_{\tau - 1}, a_{\tau - 1}) \right] + \eta \mathbf{H}\left[ \pi_\psi (a_\tau | s_\tau) \right] + \ln (a_\tau^*) \Bigr]
\end{aligned}
\end{equation}
where $\Tilde{Q}(a, s | \psi, \theta) = \pi_\psi (a_\tau | s_\tau) p_\theta (s_\tau | s_{\tau - 1}, a_{\tau-1})$. The coefficients of the generative world model's entropy $\zeta$ and the actor distribution entropy $\eta$ are set to $3 \times 10^{-4}$ in order to perform percentile exponential moving average normalization (as in~\cite{hafner2023dreamerv3}). Generally, maximizing this objective is equivalent to minimizing EFE. Note that, by default, the actor objective applies to continuous action spaces. For discrete action spaces, we may adapt the training of the policy network using the straight-through gradient estimator~\cite{Bengio2013StraightThroughGradient}, further motivated by the approach taken in~\cite{hafner2020dreamerv2}.

\section{Experimental Results}
\label{sec:exp}

We compare the performance of our agent with relevant model-based RL and AIF baselines, namely: 1)  DreamerV3~\cite{hafner2023dreamerv3}, 2) our generalization of model in~\cite{millidge2020VariationalPolicyGradients} (DAIVPG-G), and 3) the model in~\cite{vanderHimst2020AIFPOMDPCartPole} (Himst-G). Additionally, we change the architecture of the active inference agent of~\cite{vanderHimst2020AIFPOMDPCartPole} by replacing the 3D-convolution (applied over four stacked frames) with the state space model to make the agent operate properly in a POMDP environment (e.g., allowing it to process one image, instead of stacked frames, at each time-step, which we found improve the model's performance and overall stability). 
For each baseline agent and benchmark environment, we run each experiment/simulation for $4$ uniquely seeded trials, simulating each agent for $1$, $3$, and $5$ million steps on the mountain car, Meta-World~\cite{Yu2019Metaworld}, and robosuite~\cite{Zhu2020robosuite} environments, respectively. Agent performance is reported as the mean and standard deviation of the following statistics: 
\textbf{1)} average cumulative reward (ACR), 
\textbf{2)} relative stability (R-S) (proposed in~\cite{Ororbia2023ActivePredictiveCoding}, for characterizing the quality of a robotic controller's convergence ability), and 
\textbf{3)} (task) success rate (SR). 
These statistics are computed from the last $100$ recorded agent training episodes, yielding us an estimate of its online learning performance. We also train the expert (used for simulating collected imitation data) using SAC~\cite{Haarnoja2018SAC} (for robosuite) and PPO~\cite{Schulman2017PPO} (for the other problems) in the MDP version of each environment. $10,000$ steps of these collected expert data are used to produce a mean cumulative reward upper bound for the performance expected of experimental agents (see Figure~\ref{fig:reward-trends}). For the R-AIF agent's expert/positive data collection, we record about $1,000$ steps for mountain car (about $22$ episodes), $3,300$ steps for Meta-World (about $6$ episodes), and $20,000$ steps for robosuite (about $40$ episodes).

\textbf{Environment Setup.} We perform experiments on three main problem environments: mountain car~\cite{Sutton2018RLBook}, Meta-World~\cite{Yu2019Metaworld} (tasks start with ``M''), and robosuite~\cite{Zhu2020robosuite} (tasks start with ``R''). The mountain car problem is a single sparse reward task, Meta-World contains $13$ different tasks, and robosuite contains $2$ robotic control tasks. For the robotic tasks, we utilize a combination of (camera) viewpoints as the image observation, including a top-down view, an agent workspace, a front camera view, and a side view (three of these are available in Meta-World, all four in robosuite. We also modify all environments such that the reward signal is sparse: $0$ ($1$ for robosuite) is provided when the agent achieves the goal and $-1$ ($0$ for robosuite) everywhere else.

\begin{longtable}{|c||c|c|c|}
  \endhead
    \hline
		\textbf{Mountain Car} & \textbf{ACR} & \textbf{R-S} & \textbf{SR} \\
		R-AIF (Ours) & \cellcolor{LightCyan}$-68.3 \pm 1.0$ & \cellcolor{LightCyan}$0.1 \pm 0.0$ & \cellcolor{LightCyan}$1.0 \pm 0.0$ \\
		DreamerV3 & $-79.7 \pm 12.4$ & $0.4 \pm 0.2$ & $0.9 \pm 0.1$ \\
		DAIVPG-G & $-200.0 \pm 0.0$ & $0.7 \pm 0.5$ & $0.0 \pm 0.0$ \\
		Himst-G & $-199.9 \pm 0.2$ & $0.8 \pm 0.3$ & $0.0 \pm 0.0$ \\
		\hline
		\textbf{M Button Press} & \textbf{ACR} & \textbf{R-S} & \textbf{SR} \\
		R-AIF (Ours) & $-38.2 \pm 1.2$ & \cellcolor{LightCyan}$0.0 \pm 0.0$ & \cellcolor{LightCyan}$1.0 \pm 0.0$ \\
		DreamerV3 & \cellcolor{LightCyan}$-37.2 \pm 1.9$ & \cellcolor{LightCyan}$0.0 \pm 0.0$ & \cellcolor{LightCyan}$1.0 \pm 0.0$ \\
		DAIVPG-G & $-477.3 \pm 3.3$ & $0.9 \pm 0.0$ & $0.0 \pm 0.0$ \\
		Himst-G & $-365.7 \pm 163.0$ & $0.4 \pm 0.4$ & $0.3 \pm 0.4$ \\
		\hline
		\textbf{M Drawer Close} & \textbf{ACR} & \textbf{R-S} & \textbf{SR} \\
		R-AIF (Ours) & $-18.3 \pm 0.3$ & \cellcolor{LightCyan}$0.0 \pm 0.0$ & \cellcolor{LightCyan}$1.0 \pm 0.0$ \\
		DreamerV3 & \cellcolor{LightCyan}$-15.5 \pm 0.1$ & \cellcolor{LightCyan}$0.0 \pm 0.0$ & \cellcolor{LightCyan}$1.0 \pm 0.0$ \\
		DAIVPG-G & $-73.6 \pm 42.4$ & $0.1 \pm 0.1$ & \cellcolor{LightCyan}$1.0 \pm 0.1$ \\
		Himst-G & $-45.0 \pm 11.6$ & $0.3 \pm 0.4$ & $0.7 \pm 0.4$ \\
		\hline
		\textbf{M Window Open} & \textbf{ACR} & \textbf{R-S} & \textbf{SR} \\
		R-AIF (Ours) & $-48.1 \pm 2.3$ & \cellcolor{LightCyan}$0.0 \pm 0.0$ & \cellcolor{LightCyan}$1.0 \pm 0.0$ \\
		DreamerV3 & \cellcolor{LightCyan}$-44.8 \pm 2.0$ & \cellcolor{LightCyan}$0.0 \pm 0.0$ & \cellcolor{LightCyan}$1.0 \pm 0.0$ \\
		DAIVPG-G & $-354.2 \pm 69.0$ & $0.7 \pm 0.2$ & $0.4 \pm 0.2$ \\
		Himst-G & $-283.2 \pm 126.1$ & $0.5 \pm 0.3$ & $0.5 \pm 0.3$ \\
		\hline
		\textbf{M Handle Pull} & \textbf{ACR} & \textbf{R-S} & \textbf{SR} \\
		R-AIF (Ours) & \cellcolor{LightCyan}$-42.1 \pm 2.0$ & \cellcolor{LightCyan}$0.0 \pm 0.0$ & \cellcolor{LightCyan}$1.0 \pm 0.0$ \\
		DreamerV3 & $-145.4 \pm 193.8$ & $0.2 \pm 0.3$ & $0.7 \pm 0.4$ \\
		DAIVPG-G & $-416.9 \pm 72.4$ & $0.8 \pm 0.1$ & $0.3 \pm 0.3$ \\
		Himst-G & $-461.3 \pm 34.1$ & $0.9 \pm 0.1$ & $0.1 \pm 0.2$ \\
		\hline
		\textbf{M Door Close} & \textbf{ACR} & \textbf{R-S} & \textbf{SR} \\
		R-AIF (Ours) & $-51.4 \pm 0.6$ & \cellcolor{LightCyan}$0.0 \pm 0.0$ & \cellcolor{LightCyan}$1.0 \pm 0.0$ \\
		DreamerV3 & \cellcolor{LightCyan}$-50.7 \pm 0.2$ & \cellcolor{LightCyan}$0.0 \pm 0.0$ & \cellcolor{LightCyan}$1.0 \pm 0.0$ \\
		DAIVPG-G & $-161.6 \pm 66.2$ & $0.2 \pm 0.1$ & $0.9 \pm 0.1$ \\
		Himst-G & $-83.6 \pm 4.7$ & $0.1 \pm 0.0$ & \cellcolor{LightCyan}$1.0 \pm 0.0$ \\
		\hline
		\textbf{M Door Open} & \textbf{ACR} & \textbf{R-S} & \textbf{SR} \\
		R-AIF (Ours) & \cellcolor{LightCyan}$-62.5 \pm 0.6$ & \cellcolor{LightCyan}$0.0 \pm 0.0$ & \cellcolor{LightCyan}$1.0 \pm 0.0$ \\
		DreamerV3 & $-69.9 \pm 2.6$ & \cellcolor{LightCyan}$0.0 \pm 0.0$ & \cellcolor{LightCyan}$1.0 \pm 0.0$ \\
		DAIVPG-G & $-440.0 \pm 54.5$ & $0.8 \pm 0.1$ & $0.2 \pm 0.2$ \\
		Himst-G & $-481.0 \pm 0.0$ & $0.7 \pm 0.2$ & $0.0 \pm 0.0$ \\
		\hline
		\textbf{M Pick Place} & \textbf{ACR} & \textbf{R-S} & \textbf{SR} \\
		R-AIF (Ours) & \cellcolor{LightCyan}$-462.4 \pm 38.8$ & $0.9 \pm 0.1$ & \cellcolor{LightCyan}$0.1 \pm 0.1$ \\
		DreamerV3 & $-481.0 \pm 0.0$ & $0.8 \pm 0.1$ & $0.0 \pm 0.0$ \\
		DAIVPG-G & $-480.9 \pm 0.1$ & $0.9 \pm 0.0$ & $0.0 \pm 0.0$ \\
		Himst-G & $-481.0 \pm 0.0$ & \cellcolor{LightCyan}$0.3 \pm 0.4$ & $0.0 \pm 0.0$ \\
		\hline
		\textbf{M Push} & \textbf{ACR} & \textbf{R-S} & \textbf{SR} \\
		R-AIF (Ours) & \cellcolor{LightCyan}$-403.1 \pm 43.4$ & $0.8 \pm 0.1$ & \cellcolor{LightCyan}$0.6 \pm 0.3$ \\
		DreamerV3 & $-470.6 \pm 8.9$ & $0.9 \pm 0.0$ & $0.1 \pm 0.1$ \\
		DAIVPG-G & $-478.2 \pm 1.8$ & $0.9 \pm 0.0$ & $0.1 \pm 0.0$ \\
		Himst-G & $-481.0 \pm 0.0$ & \cellcolor{LightCyan}$0.6 \pm 0.4$ & $0.0 \pm 0.0$ \\
		\hline
		\textbf{M Reach} & \textbf{ACR} & \textbf{R-S} & \textbf{SR} \\
		R-AIF (Ours) & \cellcolor{LightCyan}$-24.4 \pm 1.4$ & \cellcolor{LightCyan}$0.0 \pm 0.0$ & \cellcolor{LightCyan}$1.0 \pm 0.0$ \\
		DreamerV3 & $-27.1 \pm 10.4$ & \cellcolor{LightCyan}$0.0 \pm 0.0$ & \cellcolor{LightCyan}$1.0 \pm 0.0$ \\
		DAIVPG-G & $-426.6 \pm 44.4$ & $0.8 \pm 0.1$ & $0.6 \pm 0.2$ \\
		Himst-G & $-481.0 \pm 0.1$ & $0.4 \pm 0.0$ & $0.0 \pm 0.0$ \\
		\hline
		\textbf{M Soccer} & \textbf{ACR} & \textbf{R-S} & \textbf{SR} \\
		R-AIF (Ours) & \cellcolor{LightCyan}$-332.7 \pm 57.8$ & \cellcolor{LightCyan}$0.6 \pm 0.1$ & \cellcolor{LightCyan}$0.5 \pm 0.2$ \\
		DreamerV3 & $-385.5 \pm 55.0$ & $0.8 \pm 0.1$ & $0.3 \pm 0.2$ \\
		DAIVPG-G & $-476.4 \pm 1.1$ & $1.0 \pm 0.0$ & $0.0 \pm 0.0$ \\
		Himst-G & $-479.7 \pm 1.8$ & $1.0 \pm 0.0$ & $0.0 \pm 0.0$ \\
		\hline
		\textbf{M Plate Slide} & \textbf{ACR} & \textbf{R-S} & \textbf{SR} \\
		R-AIF (Ours) & \cellcolor{LightCyan}$-38.3 \pm 7.0$ & \cellcolor{LightCyan}$0.0 \pm 0.0$ & \cellcolor{LightCyan}$1.0 \pm 0.0$ \\
		DreamerV3 & $-114.7 \pm 136.7$ & $0.2 \pm 0.3$ & $0.8 \pm 0.3$ \\
		DAIVPG-G & $-481.0 \pm 0.0$ & $0.6 \pm 0.4$ & $0.0 \pm 0.0$ \\
		Himst-G & $-481.0 \pm 0.0$ & $1.0 \pm 0.0$ & $0.0 \pm 0.0$ \\
		\hline
		\textbf{M Disassemble} & \textbf{ACR} & \textbf{R-S} & \textbf{SR} \\
		R-AIF (Ours) & \cellcolor{LightCyan}$-479.8 \pm 2.4$ & $0.6 \pm 0.3$ & \cellcolor{LightCyan}$0.0 \pm 0.1$ \\
		DreamerV3 & $-481.0 \pm 0.0$ & $0.5 \pm 0.3$ & \cellcolor{LightCyan}$0.0 \pm 0.0$ \\
		DAIVPG-G & $-480.8 \pm 0.2$ & $0.8 \pm 0.1$ & \cellcolor{LightCyan}$0.0 \pm 0.0$ \\
		Himst-G & $-481.0 \pm 0.0$ & \cellcolor{LightCyan}$0.0 \pm 0.0$ & \cellcolor{LightCyan}$0.0 \pm 0.0$ \\
		\hline
		\textbf{M Lever Pull} & \textbf{ACR} & \textbf{R-S} & \textbf{SR} \\
		R-AIF (Ours) & \cellcolor{LightCyan}$-50.6 \pm 2.5$ & \cellcolor{LightCyan}$0.0 \pm 0.0$ & \cellcolor{LightCyan}$1.0 \pm 0.0$ \\
		DreamerV3 & $-161.3 \pm 184.8$ & $0.1 \pm 0.1$ & $0.7 \pm 0.4$ \\
		DAIVPG-G & $-480.6 \pm 0.1$ & $0.6 \pm 0.1$ & $0.1 \pm 0.0$ \\
		Himst-G & $-481.0 \pm 0.0$ & $0.5 \pm 0.1$ & $0.0 \pm 0.0$ \\
		\hline
		\textbf{R Door Open} & \textbf{ACR} & \textbf{R-S} & \textbf{SR} \\
		R-AIF (Ours) & \cellcolor{LightCyan}$425.8 \pm 10.0$ & $0.1 \pm 0.0$ & \cellcolor{LightCyan}$1.0 \pm 0.0$ \\
		DreamerV3 & $0.0 \pm 0.0$ & \cellcolor{LightCyan}$0.0 \pm 0.0$ & $0.0 \pm 0.0$ \\
		DAIVPG-G & $0.0 \pm 0.0$ & \cellcolor{LightCyan}$0.0 \pm 0.0$ & $0.0 \pm 0.0$ \\
		Himst-G & $0.4 \pm 0.4$ & $0.3 \pm 0.2$ & $0.0 \pm 0.0$ \\
		\hline
		\textbf{R Block Lift} & \textbf{ACR} & \textbf{R-S} & \textbf{SR} \\
		R-AIF (Ours) & \cellcolor{LightCyan}$120.8 \pm 159.8$ & $0.4 \pm 0.2$ & \cellcolor{LightCyan}$0.4 \pm 0.4$ \\
		DreamerV3 & $0.0 \pm 0.0$ & \cellcolor{LightCyan}$0.0 \pm 0.0$ & $0.0 \pm 0.0$ \\
		DAIVPG-G & $0.0 \pm 0.0$ & $0.1 \pm 0.1$ & $0.0 \pm 0.0$ \\
		Himst-G & $0.1 \pm 0.1$ & \cellcolor{LightCyan}$0.0 \pm 0.0$ & $0.0 \pm 0.0$ \\
		\hline
  \caption{Table shows different statistics for each baseline agent under each benchmark robotic tasks.}
  \label{tab:all-stats}
\end{longtable}

\textbf{Results.} In general, our proposed R-AIF agent obtains earlier convergence than the DreamerV3, DAIVPG-G, and Himst-G models; see Figure~\ref{fig:reward-trends}. For most of the tasks, the R-AIF agent exhibits an ability to perform successfully earlier on; we hypothesize that this happens because the agent's underlying policy is trained to ``shape'' the trajectory to one that the CRSPP sub-module (prior) prefers. Furthermore, R-AIF also obtains a final cumulative reward as well as a success rate that is higher than the three baseline algorithms for most of the tasks (see Table~\ref{tab:all-stats}). Figure~\ref{fig:reward-trends} and Table~\ref{tab:all-stats} also demonstrate that the proposed R-AIF model is more stable overall, with a standard deviation for most tasks that is lower than the other three baseline models.

Specifically, for the mountain car environment, DreamerV3 struggles to optimize effectively in the earlier phases, and its performance improves only when enough successful experiences have been collected from its random exploration. For Meta-World tasks, all agents are able to achieve at least some degree of success over time. Observe that our R-AIF agent learns to excel at a task very early on (utilizing only a small portion of collected expert/imitation experiences) even in tasks where the expert struggles to succeed (e.g., M Push). For robosuite, the proposed R-AIF is the only agent that can solve the tasks successfully compared to other baselines.

\section{Conclusions}
\label{sec:conclusions}

In this work, we crafted what we called the robust active inference (R-AIF) framework, where agents are engaged in the dynamic, active perception, manipulating their environments while driven by our proposed \textit{contrastive recurrent state prior preference}. An R-AIF agent learns to take actions by utilizing a policy network that optimizes through a generalized advantage value estimated from the instrumental and epistemic signals derived from our expected free energy objective. The instrumental (goal-orienting) signal is constructed from the (sparse) reward and the dynamics of the CRSPP model while the epistemic (exploration-driving) signal is computed from the information gain and statistics of a generative world model and the policy network. Overall, we provide empirical results showing that our R-AIF achieves greater performance compared to other baselines: DreamerV3~\cite{hafner2023dreamerv3}, DAIVPG~\cite{millidge2020VariationalPolicyGradients}, and Himst's model~\cite{vanderHimst2020AIFPOMDPCartPole}. Finally, our results also demonstrate that R-AIF agents can operate well in varied goal, sparse-reward POMDP environments. Future work can consider improving the latent state space model with methods such as variational dynamics, discrete-variable autoencoders, and attention-weighting models~\cite{Becker2022VRKN, Oord2017VQVAE, Chen2018BetaTCVAE, Anonymous2023LadderVAE, Vaswani2017AttentionIsAllYouNeed}. It would also be useful to study the integration of R-AIF into physical (neuro)robotic systems, which would entail an embodied, enactive, and survival-oriented formulation of active perception and world model learning~\cite{ororbia2023mortal}.
syle{acm}
\bibliography{main}

\begin{thebibliography}{10}

\bibitem{Andrychowicz2017HER}
{\sc Andrychowicz, M., Wolski, F., Ray, A., Schneider, J., Fong, R., Welinder, P., McGrew, B., Tobin, J., Pieter~Abbeel, O., and Zaremba, W.}
\newblock Hindsight experience replay.
\newblock In {\em Advances in Neural Information Processing Systems\/} (2017), I.~Guyon, U.~V. Luxburg, S.~Bengio, H.~Wallach, R.~Fergus, S.~Vishwanathan, and R.~Garnett, Eds., vol.~30, Curran Associates, Inc.

\bibitem{Anonymous2023LadderVAE}
{\sc Anonymous}.
\newblock Causally aligned curriculum learning.
\newblock In {\em Submitted to The Twelfth International Conference on Learning Representations\/} (2023).
\newblock under review.

\bibitem{Barto2013IntrinsicMotivation}
{\sc Barto, A., Mirolli, M., and Baldassarre, G.}
\newblock Novelty or surprise?
\newblock {\em Frontiers in Psychology 4\/} (2013).

\bibitem{Becker2022VRKN}
{\sc Becker, P., and Neumann, G.}
\newblock On uncertainty in deep state space models for model-based reinforcement learning.
\newblock {\em Transactions on Machine Learning Research\/} (2022).

\bibitem{Bengio2013StraightThroughGradient}
{\sc Bengio, Y., L{\'e}onard, N., and Courville, A.~C.}
\newblock Estimating or propagating gradients through stochastic neurons for conditional computation.
\newblock {\em ArXiv abs/1308.3432\/} (2013).

\bibitem{Bishop2006MLBook}
{\sc Bishop, C.~M.}
\newblock {\em Pattern Recognition and Machine Learning (Information Science and Statistics)}.
\newblock Springer-Verlag, Berlin, Heidelberg, 2006.

\bibitem{Buesing2018RLGenerativeModel}
{\sc Buesing, L., Weber, T., Racani{\`{e}}re, S., Eslami, S. M.~A., Rezende, D.~J., Reichert, D.~P., Viola, F., Besse, F., Gregor, K., Hassabis, D., and Wierstra, D.}
\newblock Learning and querying fast generative models for reinforcement learning.
\newblock {\em CoRR abs/1802.03006\/} (2018).

\bibitem{Ccatal2019BayesianPolicyAIF}
{\sc {\c{C}}atal, O., Nauta, J., Verbelen, T., Simoens, P., and Dhoedt, B.}
\newblock Bayesian policy selection using active inference.
\newblock {\em CoRR abs/1904.08149\/} (2019).

\bibitem{Ccatal2020AIFPOMDP}
{\sc {\c{C}}atal, O., Wauthier, S., De~Boom, C., Verbelen, T., and Dhoedt, B.}
\newblock Learning generative state space models for active inference.
\newblock {\em Frontiers in Computational Neuroscience 14\/} (2020), 574372.

\bibitem{Chen2018BetaTCVAE}
{\sc Chen, T.~Q., Li, X., Grosse, R., and Duvenaud, D.}
\newblock Isolating sources of disentanglement in variational autoencoders, 2018.

\bibitem{chung2014gru}
{\sc Chung, J., Çaglar G{\"u}lçehre, Cho, K., and Bengio, Y.}
\newblock Empirical evaluation of gated recurrent neural networks on sequence modeling.
\newblock {\em ArXiv abs/1412.3555\/} (2014).

\bibitem{Deci1985IntrinsicMotivationBook}
{\sc Deci, E.~L., and Ryan, R.~M.}
\newblock {\em Intrinsic Motivation and Self-Determination in Human Behavior}.
\newblock Springer US, 1985.

\bibitem{Doerr2018PRSSM}
{\sc Doerr, A., Daniel, C., Schiegg, M., Duy, N.-T., Schaal, S., Toussaint, M., and Sebastian, T.}
\newblock Probabilistic recurrent state-space models.
\newblock In {\em Proceedings of the 35th International Conference on Machine Learning\/} (10--15 Jul 2018), J.~Dy and A.~Krause, Eds., vol.~80 of {\em Proceedings of Machine Learning Research}, PMLR, pp.~1280--1289.

\bibitem{fountas2020DAIFMC}
{\sc Fountas, Z., Sajid, N., Mediano, P., and Friston, K.}
\newblock Deep active inference agents using monte-carlo methods.
\newblock {\em Advances in neural information processing systems 33\/} (2020), 11662--11675.

\bibitem{Friston2003InferenceBrain}
{\sc Friston, K.}
\newblock Learning and inference in the brain.
\newblock {\em Neural Networks 16}, 9 (2003), 1325--1352.
\newblock Neuroinformatics.

\bibitem{Friston2013LifeAsWeKnowIt}
{\sc Friston, K.}
\newblock Life as we know it.
\newblock {\em Journal of the Royal Society, Interface / the Royal Society 10\/} (06 2013), 20130475.

\bibitem{Friston2021SophisticatedInference}
{\sc Friston, K., Da~Costa, L., Hafner, D., Hesp, C., and Parr, T.}
\newblock Sophisticated inference.
\newblock {\em Neural Computation 33}, 3 (03 2021), 713--763.

\bibitem{friston2017AIFProcessTheory}
{\sc Friston, K., FitzGerald, T., Rigoli, F., Schwartenbeck, P., and Pezzulo, G.}
\newblock Active inference: a process theory.
\newblock {\em Neural computation 29}, 1 (2017), 1--49.

\bibitem{friston2016learning}
{\sc Friston, K., FitzGerald, T., Rigoli, F., Schwartenbeck, P., Pezzulo, G., et~al.}
\newblock Active inference and learning.
\newblock {\em Neuroscience \& Biobehavioral Reviews 68\/} (2016), 862--879.

\bibitem{Friston2006FEPBrain}
{\sc Friston, K., Kilner, J., and Harrison, L.}
\newblock A free energy principle for the brain.
\newblock {\em Journal of Physiology-Paris 100}, 1 (2006), 70--87.
\newblock Theoretical and Computational Neuroscience: Understanding Brain Functions.

\bibitem{Friston2015Epistemic}
{\sc Friston, K., Rigoli, F., Ognibene, D., Mathys, C., FitzGerald, T., and Pezzulo, G.}
\newblock Active inference and epistemic value.
\newblock {\em Cognitive neuroscience\/} (02 2015).

\bibitem{Friston2010FEP}
{\sc Friston, K.~J.}
\newblock The free-energy principle: a unified brain theory?
\newblock {\em Nature Reviews Neuroscience 11\/} (2010), 127--138.

\bibitem{friston2017curiosity}
{\sc Friston, K.~J., Lin, M., Frith, C.~D., Pezzulo, G., Hobson, J.~A., and Ondobaka, S.}
\newblock {Active Inference, Curiosity and Insight}.
\newblock {\em Neural Computation 29}, 10 (10 2017), 2633--2683.

\bibitem{friston2017graphical}
{\sc Friston, K.~J., Parr, T., and de~Vries, B.}
\newblock The graphical brain: belief propagation and active inference.
\newblock {\em Network Neuroscience 1}, 4 (2017), 381--414.

\bibitem{Gal2015DropoutBayesian}
{\sc Gal, Y., and Ghahramani, Z.}
\newblock Dropout as a bayesian approximation: Representing model uncertainty in deep learning.
\newblock In {\em Proceedings of The 33rd International Conference on Machine Learning\/} (New York, New York, USA, 20--22 Jun 2016), M.~F. Balcan and K.~Q. Weinberger, Eds., vol.~48 of {\em Proceedings of Machine Learning Research}, PMLR, pp.~1050--1059.

\bibitem{Goodfellow2016DeepLearningBook}
{\sc Goodfellow, I., Bengio, Y., and Courville, A.}
\newblock {\em Deep Learning}.
\newblock MIT Press, 2016.
\newblock \url{http://www.deeplearningbook.org}.

\bibitem{Ha2018WorldModel}
{\sc Ha, D.~R., and Schmidhuber, J.}
\newblock World models.
\newblock {\em ArXiv abs/1803.10122\/} (2018).

\bibitem{Haarnoja2018SAC}
{\sc Haarnoja, T., Zhou, A., Abbeel, P., and Levine, S.}
\newblock Soft actor-critic: Off-policy maximum entropy deep reinforcement learning with a stochastic actor.
\newblock In {\em Proceedings of the 35th International Conference on Machine Learning, {ICML} 2018, Stockholmsm{\"{a}}ssan, Stockholm, Sweden, July 10-15, 2018\/} (2018), J.~G. Dy and A.~Krause, Eds., vol.~80 of {\em Proceedings of Machine Learning Research}, {PMLR}, pp.~1856--1865.

\bibitem{Haarnoja2018SACAPP}
{\sc Haarnoja, T., Zhou, A., Hartikainen, K., Tucker, G., Ha, S., Tan, J., Kumar, V., Zhu, H., Gupta, A., Abbeel, P., and Levine, S.}
\newblock Soft actor-critic algorithms and applications, 2018.

\bibitem{Hafner2020Dreamerv1}
{\sc Hafner, D., Lillicrap, T., Ba, J., and Norouzi, M.}
\newblock Dream to control: Learning behaviors by latent imagination.
\newblock In {\em International Conference on Learning Representations\/} (2020).

\bibitem{Hafner2019Planet}
{\sc Hafner, D., Lillicrap, T.~P., Fischer, I., Villegas, R., Ha, D., Lee, H., and Davidson, J.}
\newblock Learning latent dynamics for planning from pixels.
\newblock In {\em Proceedings of the 36th International Conference on Machine Learning, {ICML} 2019, 9-15 June 2019, Long Beach, California, {USA}\/} (2019), K.~Chaudhuri and R.~Salakhutdinov, Eds., vol.~97 of {\em Proceedings of Machine Learning Research}, {PMLR}, pp.~2555--2565.

\bibitem{hafner2020dreamerv2}
{\sc Hafner, D., Lillicrap, T.~P., Norouzi, M., and Ba, J.}
\newblock Mastering atari with discrete world models.
\newblock In {\em International Conference on Learning Representations\/} (2021).

\bibitem{Hafner2020ActionAP}
{\sc Hafner, D., Ortega, P.~A., Ba, J., Parr, T., Friston, K.~J., and Heess, N. M.~O.}
\newblock Action and perception as divergence minimization.
\newblock {\em ArXiv abs/2009.01791\/} (2020).

\bibitem{hafner2023dreamerv3}
{\sc Hafner, D., Pasukonis, J., Ba, J., and Lillicrap, T.}
\newblock Mastering diverse domains through world models.
\newblock {\em ArXiv abs/2301.04104\/} (2023).

\bibitem{Han2023RLRobotSurvey}
{\sc Han, D., Mulyana, B., Stankovic, V., and Cheng, S.}
\newblock A survey on deep reinforcement learning algorithms for robotic manipulation.
\newblock {\em Sensors 23}, 7 (2023).

\bibitem{he2020momentum}
{\sc He, K., Fan, H., Wu, Y., Xie, S., and Girshick, R.}
\newblock Momentum contrast for unsupervised visual representation learning.
\newblock In {\em 2020 IEEE/CVF Conference on Computer Vision and Pattern Recognition (CVPR)\/} (2020), pp.~9726--9735.

\bibitem{Hoffman2013StochasticVI}
{\sc Hoffman, M.~D., Blei, D.~M., Wang, C., and Paisley, J.}
\newblock Stochastic variational inference.
\newblock {\em Journal of Machine Learning Research 14}, 40 (2013), 1303--1347.

\bibitem{Karl2017VariationalBayesFilters}
{\sc Karl, M., Soelch, M., Bayer, J., and van~der Smagt, P.}
\newblock Deep variational bayes filters: Unsupervised learning of state space models from raw data.
\newblock In {\em International Conference on Learning Representations\/} (2017).

\bibitem{Kingma2014VAE}
{\sc Kingma, D.~P., and Welling, M.}
\newblock Auto-encoding variational bayes.
\newblock In {\em 2nd International Conference on Learning Representations, {ICLR} 2014, Banff, AB, Canada, April 14-16, 2014, Conference Track Proceedings\/} (2014).

\bibitem{Krayani2022AIFUAV}
{\sc Krayani, A., Alam, A.~S., Marcenaro, L., Nallanathan, A., and Regazzoni, C.}
\newblock A novel resource allocation for anti-jamming in cognitive-uavs: An active inference approach.
\newblock {\em IEEE Communications Letters 26}, 10 (2022), 2272--2276.

\bibitem{laskin2020curl}
{\sc Laskin, M., Srinivas, A., and Abbeel, P.}
\newblock {CURL}: Contrastive unsupervised representations for reinforcement learning.
\newblock In {\em Proceedings of the 37th International Conference on Machine Learning\/} (13--18 Jul 2020), H.~D. III and A.~Singh, Eds., vol.~119 of {\em Proceedings of Machine Learning Research}, PMLR, pp.~5639--5650.

\bibitem{Lecun2015DeepLearningBook}
{\sc LeCun, Y., Bengio, Y., and Hinton, G.}
\newblock Deep learning.
\newblock {\em Nature 521\/} (05 2015), 436--44.

\bibitem{Lindley1956IG}
{\sc Lindley, D.~V.}
\newblock {On a Measure of the Information Provided by an Experiment}.
\newblock {\em The Annals of Mathematical Statistics 27}, 4 (1956), 986 -- 1005.

\bibitem{MacKay2003InformationTheoryBook}
{\sc MacKay, D. J.~C.}
\newblock {\em Information Theory, Inference, and Learning Algorithms}.
\newblock Copyright Cambridge University Press, 2003.

\bibitem{Mazzaglia2021ContrastiveAIF}
{\sc Mazzaglia, P., Verbelen, T., and Dhoedt, B.}
\newblock Contrastive active inference.
\newblock In {\em Advances in Neural Information Processing Systems\/} (2021), A.~Beygelzimer, Y.~Dauphin, P.~Liang, and J.~W. Vaughan, Eds.

\bibitem{Mazzaglia2022DeepAIFSurvey}
{\sc Mazzaglia, P., Verbelen, T., Çatal, O., and Dhoedt, B.}
\newblock The free energy principle for perception and action: A deep learning perspective.
\newblock {\em Entropy 24}, 2 (2022).

\bibitem{millidge2020VariationalPolicyGradients}
{\sc Millidge, B.}
\newblock Deep active inference as variational policy gradients.
\newblock {\em Journal of Mathematical Psychology 96\/} (2020), 102348.

\bibitem{Millidge2021WhenceEFE}
{\sc Millidge, B., Tschantz, A., and Buckley, C.}
\newblock Whence the expected free energy?
\newblock {\em Neural Computation 33\/} (01 2021), 1--36.

\bibitem{Mnih2013DQN}
{\sc Mnih, V., Kavukcuoglu, K., Silver, D., Graves, A., Antonoglou, I., Wierstra, D., and Riedmiller, M.~A.}
\newblock Playing atari with deep reinforcement learning.
\newblock {\em CoRR abs/1312.5602\/} (2013).

\bibitem{mnih2015human}
{\sc Mnih, V., Kavukcuoglu, K., Silver, D., Rusu, A.~A., Veness, J., Bellemare, M.~G., Graves, A., Riedmiller, M., Fidjeland, A.~K., Ostrovski, G., et~al.}
\newblock Human-level control through deep reinforcement learning.
\newblock {\em nature 518}, 7540 (2015), 529--533.

\bibitem{Morales2021RLRobotSurvey}
{\sc Morales, E.~F., Murrieta-Cid, R., Becerra, I., and Esquivel-Basaldua, M.~A.}
\newblock A survey on deep learning and deep reinforcement learning in robotics with a tutorial on deep reinforcement learning.
\newblock {\em Intell. Serv. Robot. 14}, 5 (nov 2021), 773–805.

\bibitem{Noel2021AIFCapsule}
{\sc Noel, A.~D., van Hoof, C., and Millidge, B.}
\newblock Online reinforcement learning with sparse rewards through an active inference capsule.
\newblock {\em ArXiv abs/2106.02390\/} (2021).

\bibitem{Nozari2022AIFAutonomousDriving}
{\sc Nozari, S., Krayani, A., Marin-Plaza, P., Marcenaro, L., Gómez, D.~M., and Regazzoni, C.}
\newblock Active inference integrated with imitation learning for autonomous driving.
\newblock {\em IEEE Access 10\/} (2022), 49738--49756.

\bibitem{Okada2021Dreaming}
{\sc Okada, M., and Taniguchi, T.}
\newblock Dreaming: Model-based reinforcement learning by latent imagination without reconstruction.
\newblock In {\em 2021 IEEE International Conference on Robotics and Automation (ICRA)\/} (2021), pp.~4209--4215.

\bibitem{Oliver2022aifhumanrobot}
{\sc Oliver, G., Lanillos, P., and Cheng, G.}
\newblock An empirical study of active inference on a humanoid robot.
\newblock {\em IEEE Transactions on Cognitive and Developmental Systems 14}, 2 (2022), 462--471.

\bibitem{ororbia2023mortal}
{\sc Ororbia, A., and Friston, K.}
\newblock Mortal computation: A foundation for biomimetic intelligence.
\newblock {\em arXiv preprint arXiv:2311.09589\/} (2023).

\bibitem{Ororbia2023ActivePredictiveCoding}
{\sc Ororbia, A., and Mali, A.}
\newblock Active predictive coding: Brain-inspired reinforcement learning for sparse reward robotic control problems.
\newblock In {\em 2023 IEEE International Conference on Robotics and Automation (ICRA)\/} (2023), pp.~3015--3021.

\bibitem{Oudeyer2007IntrinsicMotivation}
{\sc Oudeyer, P.-Y., and Kaplan, F.}
\newblock What is intrinsic motivation? a typology of computational approaches.
\newblock {\em Frontiers in Neurorobotics 1\/} (2007).

\bibitem{parr2019generalised}
{\sc Parr, T., and Friston, K.~J.}
\newblock Generalised free energy and active inference.
\newblock {\em Biological cybernetics 113}, 5 (2019), 495--513.

\bibitem{Parr2019MarginalMessagePassing}
{\sc Parr, T., Markovi{\'c}, D., Kiebel, S.~J., and Friston, K.~J.}
\newblock Neuronal message passing using mean-field, bethe, and marginal approximations.
\newblock {\em Scientific Reports 9\/} (2019).

\bibitem{Parr2022AIFFEPBook}
{\sc Parr, T., Pezzulo, G., and Friston, K.}
\newblock {\em Active Inference: The Free Energy Principle in Mind, Brain, and Behavior}.
\newblock 01 2022.

\bibitem{pathak2017curiosity}
{\sc Pathak, D., Agrawal, P., Efros, A.~A., and Darrell, T.}
\newblock Curiosity-driven exploration by self-supervised prediction.
\newblock In {\em International conference on machine learning\/} (2017), PMLR, pp.~2778--2787.

\bibitem{Paul2023DPEFE}
{\sc Paul, A., Sajid, N., Costa, L.~D., and Razi, A.}
\newblock On efficient computation in active inference, 2023.

\bibitem{Pezzulo2024InteractWithWorld}
{\sc Pezzulo, G., D'Amato, L., Mannella, F., Priorelli, M., Van~de Maele, T., Stoianov, I.~P., and Friston, K.}
\newblock Neural representation in active inference: Using generative models to interact with—and understand—the lived world.
\newblock {\em Annals of the New York Academy of Sciences 1534}, 1 (2024), 45--68.

\bibitem{Racaniere2017ModelBasedRLAugmented}
{\sc Racani\`{e}re, S., Weber, T., Reichert, D., Buesing, L., Guez, A., Jimenez~Rezende, D., Puigdom\`{e}nech~Badia, A., Vinyals, O., Heess, N., Li, Y., Pascanu, R., Battaglia, P., Hassabis, D., Silver, D., and Wierstra, D.}
\newblock Imagination-augmented agents for deep reinforcement learning.
\newblock In {\em Advances in Neural Information Processing Systems\/} (2017), I.~Guyon, U.~V. Luxburg, S.~Bengio, H.~Wallach, R.~Fergus, S.~Vishwanathan, and R.~Garnett, Eds., vol.~30, Curran Associates, Inc.

\bibitem{Rajeswar2023UnsupervisedRLPixels}
{\sc Rajeswar, S., Mazzaglia, P., Verbelen, T., Piché, A., Dhoedt, B., Courville, A., and Lacoste, A.}
\newblock Mastering the unsupervised reinforcement learning benchmark from pixels.
\newblock In {\em 40th International Conference on Machine Learning\/} (2023).

\bibitem{Russell2010AIModernBook}
{\sc Russell, S., and Norvig, P.}
\newblock {\em Artificial Intelligence: A Modern Approach}, 3~ed.
\newblock Prentice Hall, 2010.

\bibitem{Sajid2021AIFDemystified}
{\sc Sajid, N., Ball, P.~J., Parr, T., and Friston, K.~J.}
\newblock Active inference: demystified and compared.
\newblock {\em Neural computation 33}, 3 (2021), 674--712.

\bibitem{Schulman2015TRPO}
{\sc Schulman, J., Levine, S., Abbeel, P., Jordan, M., and Moritz, P.}
\newblock Trust region policy optimization.
\newblock In {\em Proceedings of the 32nd International Conference on Machine Learning\/} (Lille, France, 07--09 Jul 2015), F.~Bach and D.~Blei, Eds., vol.~37 of {\em Proceedings of Machine Learning Research}, PMLR, pp.~1889--1897.

\bibitem{Schulman2015GeneralGAELambda}
{\sc Schulman, J., Moritz, P., Levine, S., Jordan, M.~I., and Abbeel, P.}
\newblock High-dimensional continuous control using generalized advantage estimation.
\newblock {\em CoRR abs/1506.02438\/} (2015).

\bibitem{Schulman2017PPO}
{\sc Schulman, J., Wolski, F., Dhariwal, P., Radford, A., and Klimov, O.}
\newblock Proximal policy optimization algorithms.
\newblock {\em CoRR abs/1707.06347\/} (2017).

\bibitem{Schwartenbeck2019Curiosity}
{\sc Schwartenbeck, P., Passecker, J., Hauser, T.~U., FitzGerald, T.~H., Kronbichler, M., and Friston, K.~J.}
\newblock Computational mechanisms of curiosity and goal-directed exploration.
\newblock {\em eLife 8\/} (may 2019), e41703.

\bibitem{SekarPlan2Explore}
{\sc Sekar, R., Rybkin, O., Daniilidis, K., Abbeel, P., Hafner, D., and Pathak, D.}
\newblock Planning to explore via self-supervised world models.
\newblock In {\em Proceedings of the 37th International Conference on Machine Learning\/} (13--18 Jul 2020), H.~D. III and A.~Singh, Eds., vol.~119 of {\em Proceedings of Machine Learning Research}, PMLR, pp.~8583--8592.

\bibitem{Shyam2018MAX}
{\sc Shyam, P., Jaśkowski, W., and Gomez, F.~J.}
\newblock Model-based active exploration.
\newblock In {\em International Conference on Machine Learning\/} (2018).

\bibitem{Smith2022TutorialPaper}
{\sc Smith, R., Friston, K.~J., and Whyte, C.~J.}
\newblock A step-by-step tutorial on active inference and its application to empirical data.
\newblock {\em Journal of Mathematical Psychology 107\/} (2022), 102632.

\bibitem{Sutton1991Dyna}
{\sc Sutton, R.~S.}
\newblock Dyna, an integrated architecture for learning, planning, and reacting.
\newblock {\em SIGART Bull. 2}, 4 (jul 1991), 160–163.

\bibitem{Sutton2018RLBook}
{\sc Sutton, R.~S., and Barto, A.~G.}
\newblock {\em Reinforcement Learning: An Introduction}, second~ed.
\newblock The MIT Press, 2018.

\bibitem{Tschantz2020Scaling}
{\sc Tschantz, A., Baltieri, M., Seth, A.~K., and Buckley, C.~L.}
\newblock Scaling active inference.
\newblock In {\em 2020 International Joint Conference on Neural Networks (IJCNN)\/} (2020), pp.~1--8.

\bibitem{tschantz2020reinforcement}
{\sc Tschantz, A., Millidge, B., Seth, A.~K., and Buckley, C.~L.}
\newblock Reinforcement learning through active inference.
\newblock {\em CoRR abs/2002.12636\/} (2020).

\bibitem{Oord2018CPC}
{\sc van~den Oord, A., Li, Y., and Vinyals, O.}
\newblock Representation learning with contrastive predictive coding.
\newblock {\em CoRR abs/1807.03748\/} (2018).

\bibitem{Oord2017VQVAE}
{\sc van~den Oord, A., Vinyals, O., and Kavukcuoglu, K.}
\newblock Neural discrete representation learning.
\newblock In {\em Proceedings of the 31st International Conference on Neural Information Processing Systems\/} (Red Hook, NY, USA, 2017), NIPS'17, Curran Associates Inc., p.~6309–6318.

\bibitem{vanderHimst2020AIFPOMDPCartPole}
{\sc van~der Himst, O., and Lanillos, P.}
\newblock Deep active inference for partially observable mdps.
\newblock In {\em Active Inference\/} (Cham, 2020), T.~Verbelen, P.~Lanillos, C.~L. Buckley, and C.~De~Boom, Eds., Springer International Publishing, pp.~61--71.

\bibitem{Hoeffelen2021DAICarRacing}
{\sc van Hoeffelen, N., and Lanillos, P.}
\newblock Deep active inference for pixel-based discrete control: Evaluation on the car racing problem.
\newblock In {\em Machine Learning and Principles and Practice of Knowledge Discovery in Databases - International Workshops of ECML PKDD 2021, Virtual Event, September 13-17, 2021, Proceedings, Part I\/} (2021), vol.~1524 of {\em Communications in Computer and Information Science}, Springer, pp.~843--856.

\bibitem{Vaswani2017AttentionIsAllYouNeed}
{\sc Vaswani, A., Shazeer, N., Parmar, N., Uszkoreit, J., Jones, L., Gomez, A.~N., Kaiser, L.~u., and Polosukhin, I.}
\newblock Attention is all you need.
\newblock In {\em Advances in Neural Information Processing Systems\/} (2017), I.~Guyon, U.~V. Luxburg, S.~Bengio, H.~Wallach, R.~Fergus, S.~Vishwanathan, and R.~Garnett, Eds., vol.~30, Curran Associates, Inc.

\bibitem{Watter2015E2C}
{\sc Watter, M., Springenberg, J., Boedecker, J., and Riedmiller, M.}
\newblock Embed to control: A locally linear latent dynamics model for control from raw images.
\newblock In {\em Advances in Neural Information Processing Systems\/} (2015), C.~Cortes, N.~Lawrence, D.~Lee, M.~Sugiyama, and R.~Garnett, Eds., vol.~28, Curran Associates, Inc.

\bibitem{yang2023neural}
{\sc Yang, Z., Diaz, G.~J., Fajen, B.~R., Bailey, R., and Ororbia, A.~G.}
\newblock A neural active inference model of perceptual-motor learning.
\newblock {\em Frontiers in Computational Neuroscience 17\/} (2023), 1099593.

\bibitem{Yu2019Metaworld}
{\sc Yu, T., Quillen, D., He, Z., Julian, R., Hausman, K., Finn, C., and Levine, S.}
\newblock Meta-world: A benchmark and evaluation for multi-task and meta reinforcement learning.
\newblock In {\em Conference on Robot Learning (CoRL)\/} (2019).

\bibitem{Zhu2020robosuite}
{\sc Zhu, Y., Wong, J., Mandlekar, A., Mart\'{i}n-Mart\'{i}n, R., Joshi, A., Nasiriany, S., and Zhu, Y.}
\newblock robosuite: A modular simulation framework and benchmark for robot learning.
\newblock In {\em arXiv preprint arXiv:2009.12293\/} (2020).

\bibitem{Kaiser2020ModelBasedRLAtari}
{\sc Łukasz Kaiser, Babaeizadeh, M., Miłos, P., Osiński, B., Campbell, R.~H., Czechowski, K., Erhan, D., Finn, C., Kozakowski, P., Levine, S., Mohiuddin, A., Sepassi, R., Tucker, G., and Michalewski, H.}
\newblock Model based reinforcement learning for atari.
\newblock In {\em International Conference on Learning Representations\/} (2020).

\end{thebibliography}

\newpage

\section*{}
\section*{Appendix: Implementation Details Documentation}
\addcontentsline{toc}{section}{Detailed Implementation Documentation}
\renewcommand{\thesubsection}{\Alph{subsection}}

\subsection{Recurrent State Space Model and World Model}

\textbf{Temporal Information.} In active inference, the hidden state inferred by the agent is often computed by a likelihood matrix $\mathbb{R}^{m \times n}$ where $m$ is the number of possible state values and $n$ is the number of possible observation values~\cite{friston2017AIFProcessTheory}. A single observation from the environment can then be directly mapped to a state using this scheme. Similarly, in the amortized inference context, the recognition density parameterized by an artificial neural network (ANN) is often used to estimate the posterior probability density over the hidden states of the environment~\cite{fountas2020DAIFMC}. However, in the POMDP setting, an observation from a single time step would not provide sufficient information about the state as is done in classic active inference literature with a likelihood matrix. For example, higher-order information, such as velocity and acceleration of particular variables, cannot be captured in one single image but instead must be inferred from a sequence of images (or manually integrated~\cite{parr2019generalised}). Therefore, it is crucial for every active inference model in POMDP environments to maintain temporal information or deterministic beliefs throughout time as additional information is input into the generative model framework.

\begin{figure}
    \centering
    \includegraphics[width=0.8\linewidth]{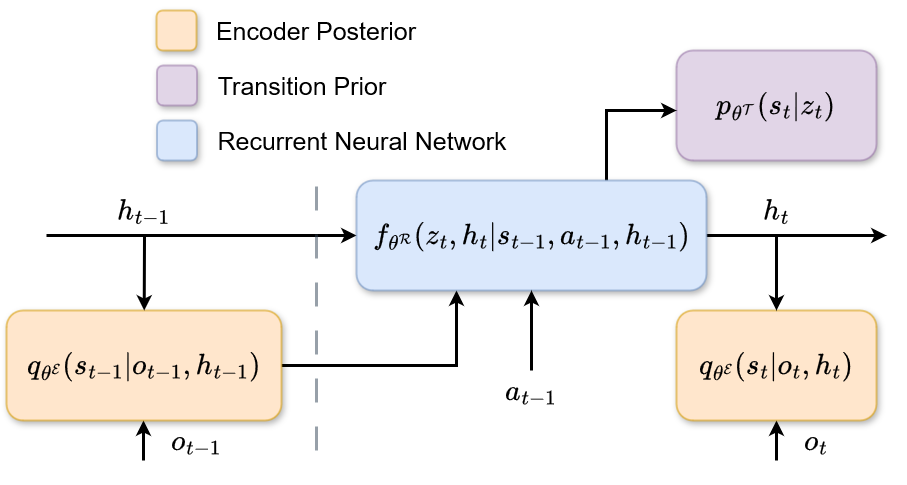}
    \caption{\textbf{The temporal generative dynamics model.} A depiction of the generative model that R-AIF uses to make use of past information; this is equivalent to a RSSM operating in latent space.}\label{fig:rssm}
\end{figure}

Since active inference posits that an agent finds a policy, i.e, a sequence of actions, based on the estimated future state distribution~\cite{Friston2010FEP, Smith2022TutorialPaper, Friston2015Epistemic, Friston2013LifeAsWeKnowIt, Millidge2021WhenceEFE}, one approach is to predict the next state $s_{t+1}$ given the current state $s_t$ and action $a_t$. When operating in POMDP environments, this methodology involves predicting the next partial observation $o_{t+1}$ given the previous partial observation $o_t$ and the action $a_t$. In order to integrate temporal information into this generative model, we can use the `carried-over' recurrent state in some forms of RNN such as a gated recurrent unit~\cite{chung2014gru} as was done in~\cite{Hafner2020Dreamerv1, hafner2020dreamerv2, hafner2023dreamerv3, Noel2021AIFCapsule}. Therefore, the prior $p(s)$ and posterior $q(s)$ distributions over states are able to encapsulate the temporal information $h$ embodied in previous observations
; see Figure~\ref{fig:rssm} for a visual representation of this process.

\subsection{Improving Numerical Stability}

Minimizing the complexity (term) aids the agent in closing the gap between its prior and its approximate posterior whereas minimizing the accuracy (term) improves the model's future observation estimation. We utilize the world model which has a discretized state space~\cite{hafner2020dreamerv2} where each hidden state is represented by a vector of discrete distributions instead of a vector of Gaussian distribution parameters as is done in other deep active inference formulations. We also employ the KL balancing~\cite{hafner2020dreamerv2} and applying ``symlog'' function to inputs~\cite{hafner2023dreamerv3} for numerical stability:
\begin{equation}
\begin{aligned}
    \text{D}_{\text{KL}}\left[ q_{\theta}(s_t | o_t) \parallel p_{\theta}(s_t | s_{t-1}, a_{t-1}) \right] \gets \ &\eta_\text{rep} \text{D}_{\text{KL}}\left[ q_{\theta}(s_t | o_t) \parallel \text{sg}( p_{\theta}(s_t | s_{t-1}, a_{t-1}) ) \right] \\
    &+ \eta_\text{dyn} \text{D}_{\text{KL}}\left[ \text{sg} (q_{\theta}(s_t | o_t) ) \parallel p_{\theta}(s_t | s_{t-1}, a_{t-1}) \right]
\end{aligned}
\end{equation}
where $\text{sg}$ is the stop gradient operation, $\eta_\text{rep}$ and $\eta_\text{dyn}$ are the coefficients for representation and dynamics KL losses, respectively. We also clip the KL term to a minimum value of free bits~\cite{Hafner2020Dreamerv1} and finally apply the $symlog$ function~\cite{hafner2023dreamerv3} on its inputs for numerical stability.

\subsection{The prior preference self-revision mechanism}

For each step in a trajectory $e$, we record the following statistics: 
\textbf{1)} a boolean showing whether the step was carried out by an expert $p_t$ or not, 
\textbf{2)} a boolean representing whether the agent immediately achieved the goal at a step $d_t$, and 
\textbf{3)} a boolean indicating whether the agent succeeded in reaching its goal at least one time within the episode $k$. 
For each time step, we want to craft a decaying signal that emphasizes the degree of preference $\rho_t \in \left[-1, 1\right]$. In order to do so, we equip the agent with a \textit{self-revision} mechanism which allows the agent to ``look back'' on how it performed and determine whether a certain state is preferred or not (see Algorithm~\ref{alg:self-revision}). Note that we complete the for-loop for positive samples and set the preference at its highest value at the state and decay this backward whenever there is a successful state. In contrast, when the episode fails, the agent only needs to decay negatively backward from the end of the episode.

\subsection{Computing the Information Gain using a Network Ensemble}

Taking actions that reduce the uncertainty of model parameters requires estimating the uncertainty in the first place. One can compute such a term from an explicit ANN ensemble or by using Monte Carlo dropout~\cite{Gal2015DropoutBayesian} to compute information gain~\cite{fountas2020DAIFMC}; however, as the world model grows in size/complexity, it becomes impractical to maintain a collection of multiple world models or to sample from a large state distribution. Therefore, we instead construct a separate ensemble of small multi-layer perceptrons (MLPs)~\cite{Lecun2015DeepLearningBook, Goodfellow2016DeepLearningBook} to estimate the next state based on the current state and action -- this is the ``information gain network ensemble.'' Formally, we learn a collection of $N$ transition prior models $\{ p_{\omega_i}(s_t | s_{t-1}, a_{t-1}) \}_{i=1}^N$. We next optimize the network ensemble by minimizing the negative log likelihood between the predicted state distribution (Gaussian) and the actual state produced by the world model. In essence, we engage in maximum likelihood estimation and update ensemble parameters by maximizing Gaussian likelihood for a predicted state Gaussian with $\mu_{\omega_i}, \sigma_{\omega_i} \in \mathbb{R}^D$ and an observed next state $s_\theta$ from the world model.
This can be formally stated as: 
\begin{equation}\label{eqn:ig-mle}
\begin{aligned}
    \operatorname*{arg\,min}_{\omega_i} \mathcal{L}(\omega_i) = - \left( - \frac{D}{2} \ln (2\pi\sigma_{\omega_i}^2) - \sum_{j=1}^{D}{\left( \frac{(s_{\theta, j} - \mu_{\omega_i})^2}{2\sigma_{\omega_i}^2} \right)} \right).
\end{aligned}
\end{equation}
When training the actor $\pi_\psi$ and the value network $f_\chi$, we may compute the information gain by computing the difference between the entropy of the mixture-average of the underlying Gaussian and the average of the entropy of all Gaussian outputs from the information gain network ensemble at each rolled-out step $\tau$ in imagination space. This is also the main reason why a collection of models are used instead of a single one as was done in~\cite{Shyam2018MAX, tschantz2020reinforcement}. Given the Gaussian entropy formula $\text{ent}(\sigma) = \frac{\ln(2\pi\sigma^2)}{2} + \frac{1}{2}$, we formulate the information gain (or the uncertainty associated with model parameters) estimator in the following way:
\begin{equation}\label{eqn:IG}
\begin{aligned}
    \text{IG} = \text{ent} \left( \text{std} \left( \{ s_i \}_{i=1}^N \right) \right) - \mathbb{E}_{i} \left[ \text{ent}(\sigma_{\omega_i}) \right], s_i \sim \mathcal{N}(\mu_{\omega_i}, \sigma_{\omega_i}).
\end{aligned}
\end{equation}

\begin{algorithm}
\caption{The Prior \textit{Self-Revision} Mechanism (per Episode)} \label{alg:self-revision}
\begin{minipage}[t]{0.59\textwidth}
\begin{algorithmic}
    \State Initialize positive and negative preference rate arrays $m^+$, $m^-$.
    \State $\{p_t\}_{i=1}^{T} \gets \{p_t \odot k\}_{i=1}^{T}$ \Comment{Ensure good expert actions}
    \State // Compute positive preference rate
    \State $c \gets \text{max}(p_{T}, 0)$ \Comment{Initialize carry variable}
    \For{$t \in \{T..1\}$}
        \State $c \gets \text{max}(c \odot \alpha, d_t)$ \Comment{Discount positive rate}
        \State $c \gets c \odot (c \ge \epsilon)$ \Comment{Quantize to $0$ if smaller than $\epsilon$}
        \State $m^+ \gets \{c\} \cup m^+$ \Comment{Append to front}
    \EndFor
    \State $m^+ \gets \{\text{clip}\bigl(p + m^+_t, 0, 1\bigr)\}_{t=1}^{T}$ \Comment{Value expert states more}
    \State // Compute negative preference rate
    \State $c \gets -1 / \beta$ \Comment{Initialize carry variable}
    \For{$t \in \{T..1\}$}
        \State $c \gets c \odot \beta$ \Comment{Decay backward from the episode end}
        \State $m^- \gets \{c\} \cup m^-$ \Comment{Append to front}
    \EndFor
    \State $m^- \gets \{m^-_t \odot (1 - k)\}_{t=1}^{T}$ \Comment{No negative rate when successful}
    \State $m^- \gets \{ m^-_t \odot (1 - p_t) \}_{t=1}^T$ \Comment{No negative rate in expert steps}
    \State $\rho \gets \{0\} \cup \{\text{clip}\bigl( m^-_t + m^+_t, 0, 1 \bigr)\}_{t=2}^T$ \Comment{First state is neutral}
    \State \Return $\rho$
\end{algorithmic}
\end{minipage}
\hfill
\begin{minipage}[t]{0.39\textwidth}
\begin{algorithmic}
    \State \textbf{Prepare episode data:}
    \State Successful at least once boolean $k$.
    \State Successful step boolean $d_t$.
    \State Step is taken by the expert boolean $p_t$.
    \State Episode length $T$.
    \State
    \State \textbf{Hyperparameters:}
    \State Successful discount rate $\alpha$.
    \State Failure decay rate $\beta$.
    \State Positive signal quantization threshold $\epsilon$.
\end{algorithmic}
\end{minipage}
\end{algorithm}

\subsection{Using Percentage Exponential Moving Average (PEMA)}

We introduce and set the coefficient of the generative world model entropy $\zeta$ and actor distribution entropy $\eta = 3 \times 10^{-4}$ and perform percentile exponential moving average normalization as in~\cite{hafner2023dreamerv3}; these coefficients depend on both the reward value~\cite{hafner2023dreamerv3} and model parameters, and therefore, given that they would be impractical to dynamically adjust, we choose to keep $\zeta$ and $\eta$ fixed and small enough for numerical stability. As a result, normalizing the return to the range between $0$ and $1$ can align with $\zeta$ and $\eta$ range~\cite{hafner2023dreamerv3}. We then divide the difference between the return and the computed value by the range $S$ as in~\cite{hafner2023dreamerv3}
\begin{align}
    \text{PEMA}(\mathcal{G}_\tau, f_\chi(v_\tau | s_\tau)) = \bigl( \mathcal{G}_\tau - f_\chi(v_\tau | s_\tau) \bigr) / \max\bigl( 1, S  \bigr),
\end{align}
where $S$ is computed using percentile (Per) exponential moving average (EMA)~\cite{hafner2023dreamerv3}:
\begin{align}
    S = \text{EMA} ( \text{Per}(\mathcal{G}_\tau, 95) - \text{Per}(\mathcal{G}_\tau, 5) , 99 ).
\end{align}

\subsection{R-AIF Agent Algorithm}

\begin{algorithm}
\caption{R-AIF Agent Algorithm} \label{alg:raif}
\begin{minipage}[t]{0.64\textwidth}
\begin{algorithmic}
    \State Initialize $\theta, \phi, \psi, \chi, \{ \omega_i \}_{i=1}^N$
    \State Initialize $E$, $\mathcal{M}$, and $\mathcal{M}^+$
    \State Collect a small number of successful trajectories for $\mathcal{M}^+$
    \While{total environment steps $<$ max environment total steps}
        \For{environment step $t$ in $1..T$}
            \State $s_t \sim q_\theta(s_t | s_{t-1}, a_{t-1}, o_t)$
            \State $a_t \sim \pi_\psi (a_t | s_t)$
            \If{done}
                \State Compute $\rho$ from Algorithm~\ref{alg:self-revision}
                \State $\mathcal{M} \leftarrow \mathcal{M} \cup \{ (o_t, a_t, r_t,  \rho_t) \}_{t=1}^{T}$
                \If{successful at least once step}
                    \State $\mathcal{M}^+ \leftarrow \mathcal{M}^+ \cup \{ (o_t, a_t, r_t, \rho_t) \}_{t=1}^{T}$
                \EndIf
            \EndIf
        \EndFor
        \For{gradient step in $1..S$}
            \If{$i \mod 2 = 0$}
                \State $\{ (o_t, a_t, r_t, \rho_t) \}_{t}^{t + L} \sim \mathcal{M}$ \Comment{Draw normal buffer}
            \Else
                \State $\{ (o_t, a_t, r_t, \rho_t) \}_{t}^{t + L} \sim \mathcal{M}^+$ \Comment{Draw positive buffer}
            \EndIf
            \State $\theta \gets \theta - \xi \nabla_\theta \mathbb{E} \bigl[ \mathcal{L}_t (\theta) \bigr]$ 
            \State $\phi \gets \phi - \xi \nabla_\phi \mathbb{E} \bigl[ \mathcal{L}_t (\phi) \bigr]$ 
            \State $\{ \omega_i \}_{i=1}^N \gets \{ \omega_i - \xi \nabla_{\omega_i} \mathbb{E} \bigl[ \mathcal{L}_t (\omega_i) \bigr] \}_{i=1}^N$ 
            \State Imagine trajectories $\{ (s_\tau, a_\tau) \}_{\tau=t}^H$ from each $s_t$
            \State Imagine prior preference $\{ \hat{s}_\tau \}_{\tau=t}^H$ from each $s_t$
            \State Predict $q_\theta(r_\tau|s_\tau), f_\chi(v_\tau|s_\tau)$ 
            \State Compute information gain $\text{IG}_\tau$ 
            \State Compute target advantage value $\mathcal{G}_\tau$ 
            \State $\chi \gets \chi - \xi \nabla_\chi \mathbb{E} \bigl[ \mathcal{L}_\tau(\chi) \bigr]$ 
            \State $\psi \gets \psi + \xi \nabla_\psi \mathbb{E} \bigl[ \mathcal{L}_\tau(\psi) \bigr]$ 
        \EndFor
    \EndWhile
\end{algorithmic}
\end{minipage}
\hfill
\begin{minipage}[t]{0.35\textwidth}
\begin{algorithmic}
    \State \textbf{Prepare models' parameters:}
    \State Generative world model $\theta$.
    \State CRSPP model $\phi$.
    \State Policy network $\psi$.
    \State Value function $\chi$.
    \State Information gain ensemble $\{ \omega_i \}_{i=1}^N$.
    \State
    \State \textbf{Prepare other components:}
    \State Environment $E$.
    \State Memory buffer $\mathcal{M}$.
    \State Positive memory buffer $\mathcal{M}^+$.
    \State
    \State \textbf{Hyperparameters:}
    \State Gradient update steps $S$.
    \State Imagination horizon $H$.
    \State Batch size $B$.
    \State Sequence length $L$.
    \State Learning rate $\xi$.
    \State Ensemble size $N$.
\end{algorithmic}
\end{minipage}
\end{algorithm}

See implementation details in Algorithm~\ref{alg:raif}.

\subsection{Implementation of environments}

\textbf{Pixel-level Mountain Car.} The mountain car environment~\cite{Sutton2018RLBook} is a standard testing problem in reinforcement learning in which an agent (the car) has to drive uphill. A difficult aspect of this problem is that the gravitational force is greater than full force that can be exerted by the car such that it cannot go uphill by simply moving forward. The agent has to learn to build up the car's potential energy by driving to the opposite hill (behind it) in order to create enough acceleration to reach the goal state in front of it. The original mountain car problem was proposed as an MDP where the environment state included the exact position and velocity of the car at any step in time. In the POMDP extension of the task, agents are not permitted to use this state directly. Instead, an agent must use rendered pictures (height $64$, width $64$) as its observations (and must infer useful internal states that aid it in its completion of the taks). Critically, the reward signal provided by this task is very sparse, i.e., it is $-1$ for every step and $0$ when the agent reaches the goal, and the action space is continuous. In the actual environment, we modify it slightly to provide a dark background and light objects, e.g., white car, to improve visualization for human experimenters.

\textbf{Meta-World.} In this environment, the agent (a controllable robotic arm) has to control the proprioceptive joints (represented as a vector of continuous actions) in a velocity-based control system~\cite{Yu2019Metaworld}. Since the workspace is 3D, an observation from a single viewpoint might cause the agent to struggle when inferring environment hidden states, e.g., the height of the goal can never be inferred from the top-down view. Therefore, we construct a raw pixel observation image using three different (camera) viewpoints: these include a top-down, an agent workspace, and a side camera view. Furthermore, we ensure that the reward space for the tasks in this environment are sparse, similar in form to the sparse signals produced by the mountain car environment with a reward of $0$ provided when the agent achieves the control objective (it reaches a successful state) and $-1$ everywhere else. Note that, unlike the mountain car, the environment simulation continues even after the agent reaches the goal state.

\textbf{robosuite.} Similar to the Meta-World environment, robosuite~\cite{Zhu2020robosuite} simulates a robotics environment where the agent controls different joints as continuous actions. Since the robot's workspace in robosuite is larger, we utilized four different camera viewpoints as streams of pixel observations for the agent instead of three as we did in Meta-World, namely: bird's-eye view (similar to the top-down view in Meta-World), agent view (the agent's perspective of the workspace), side view, and front view. Additionally, we modified the environment to have a sparse reward system, yielding a reward of $1$ when the agent successfully achieves the task goal, and $0$ otherwise. Finally, the environment continues even after the agent successfully reaches the goal state.

\subsection{Training R-AIF Agent}

In contrast to on-policy learning algorithms~\cite{Schulman2015TRPO, Schulman2017PPO}, we train our agent in an off-policy fashion (using memory buffers). Specifically, we utilize two replay buffers: a standard replay buffer $\mathcal{M}$ stores all agent's encountered transitions, and a positive replay buffer $\mathcal{M}^+$ only contains episodes with at least one step successfully-reached goal state. While training, we sample from these buffers equally as training with more successful samples is found to boost the convergence of CRSPP. We then simulate the agent in the environment and train the world model, CRSPP, information gain ensemble~\cite{tschantz2020reinforcement}, policy network, and value function periodically (see Algorithm~\ref{alg:raif} for specific details).

\subsection{Derivation of Expected Free Energy}

According to \cite{Schwartenbeck2019Curiosity, fountas2020DAIFMC, friston2017curiosity}, the expected free energy given the planned action distribution $\pi$ can be formulated as:

\begin{equation}
\begin{aligned}
    G_\tau(\pi) = \mathbb{E}_{\Tilde{Q}} [ \ln Q(s_\tau, \theta | \pi) - \ln Q(o_\tau, s_\tau, \theta | \pi ) ]
\end{aligned}
\end{equation}

with $\Tilde{Q} = Q(o_\tau, s_\tau, \theta | \pi)$ the joint probability of future observation, state, and model parameter given planned action $\pi$. This expected free energy formula can be further decomposed into:

\begin{equation}
\begin{aligned}
    G(\pi, \tau) = \mathbb{E}_{\Tilde{Q}} [ \ln Q(\theta | \pi) - \ln Q(\theta | o_\tau, s_\tau, \pi) ] + \mathbb{E}_{\Tilde{Q}} [ \ln Q(s_\tau | \theta, \pi) - \ln Q(s_\tau | o_\tau, \pi) ] - \mathbb{E}_{\Tilde{Q}} [ \ln P(o_\tau) ]
\end{aligned}
\end{equation}

with the first term is the model parameter exploration, denoting the mutual information between the model parameter before and after making an observation and state. The second term is the mutual information of agent's hidden state before and after making a new observation. The third term is realizing preference term where the agent tries to compute the amount of information about the observation that matches its prior preference.

For ease of computation, we decompose each terms into a form that could be implemented practically by our active inference agent. Firstly, we can further decompose the parameter exploration term:
\begin{equation}
\begin{aligned}
    \mathbb{E}_{\Tilde{Q}} [ \ln Q(\theta | \pi) - \ln Q(\theta | o_\tau, s_\tau, \pi) ] &= \mathbb{E}_{\Tilde{Q}} [ \ln Q(\theta | \pi) + \ln Q(\pi | o_\tau, s_\tau) + \ln Q(o_\tau, s_\tau) - \ln Q(\theta, \pi,  o_\tau, s_\tau) ]\\
    &= \mathbb{E}_{\Tilde{Q}} [ \ln Q(\pi | o_\tau, s_\tau) + \ln Q(o_\tau, s_\tau) - \ln Q(o_\tau, s_\tau | \theta, \pi) - \ln Q(\pi) ]\\
    &= \mathbb{E}_{\Tilde{Q}} [ \ln Q(\pi | o_\tau, s_\tau) - \ln Q(\pi) ] + \mathbb{E}_{\Tilde{Q}} [ \ln Q(o_\tau, s_\tau) - \ln Q(o_\tau, s_\tau | \theta, \pi)  ]\\
    &= \mathbb{E}_{\Tilde{Q}} [ \ln Q(\pi | o_\tau, s_\tau) - \ln Q(\pi) ] \\
    &\ \ \ + \mathbb{E}_{\Tilde{Q}} [ \ln Q(o_\tau | s_\tau) - \ln Q(o_\tau | s_\tau, \theta, \pi) ] \\
    &\ \ \ + \mathbb{E}_{\Tilde{Q}} [ \ln Q(s_\tau) - \ln Q(s_\tau | \theta, \pi) ]\\
    &= \mathbb{E}_{Q(\theta | o_\tau, s_\tau, \pi) Q(o_\tau, s_\tau | \pi) } [\ln Q(\pi | o_\tau, s_\tau) -  \mathbb{E}_{Q(o_\tau, s_\tau)} \ln Q(\pi | o_\tau, s_\tau)]\\
    &\ \ \ + \mathbb{E}_{Q(o_\tau | s_\tau, \theta, \pi) Q(s_\tau | \theta, \pi)Q(\theta|\pi)} [\mathbb{E}_{Q(\theta, \pi)} \ln Q(o_\tau | s_\tau, \theta, \pi) - \ln Q(o_\tau | s_\tau, \theta, \pi)] \\
    &\ \ \ + \mathbb{E}_{Q(o_\tau | s_\tau, \theta, \pi) Q(s_\tau | \theta, \pi)Q(\theta| \pi)} [ \mathbb{E}_{Q(\theta, \pi)} \ln Q(s_\tau | \theta, \pi) - \ln Q(s_\tau | \theta, \pi)]\\
    &= \mathbb{E}_{Q(\theta | o_\tau, s_\tau, \pi) } \mathbf{H} [Q(\pi | o_\tau, s_\tau) ] - \mathbf{H} [ \mathbb{E}_{Q(o_\tau, s_\tau)} Q(\pi | o_\tau, s_\tau) ]\\
    &\ \ \ + \mathbb{E}_{Q(s_\tau | \theta, \pi)} \mathbf{H} [\mathbb{E}_{Q(\theta,  \pi)} Q(o_\tau | s_\tau, \theta, \pi)] - \mathbb{E}_{Q(s_\tau | \theta, \pi) Q(\theta|\pi) } \mathbf{H} [ Q (o_\tau | s_\tau, \theta, \pi)] \\ 
    &\ \ \ + \mathbf{H} [ \mathbb{E}_{Q(\theta , \pi)} Q(s_\tau | \theta, \pi)] - \mathbb{E}_{Q(\theta|\pi)} \mathbf{H} [ Q (s_\tau | \theta, \pi)]\\
    &\approx \mathbb{E}_{Q(\theta | o_\tau, s_\tau, \pi) } \mathbf{H} [Q(\pi | o_\tau, s_\tau) ]  - \mathbf{H} [ \mathbb{E}_{Q(o_\tau, s_\tau)} Q(\pi | o_\tau, s_\tau) ]\\
    &\ \ \ + \mathbf{H} [ \mathbb{E}_{Q(\theta , \pi)} Q(s_\tau | \theta, \pi)] - \mathbb{E}_{Q(\theta|\pi)} \mathbf{H} [ Q (s_\tau | \theta, \pi)]
\end{aligned}
\end{equation}
where the term $\mathbb{E}_{Q(\theta | o_\tau, s_\tau, \pi) } \mathbf{H} [\ln Q(\pi | o_\tau, s_\tau) ]$ is defined as entropy of the predicted action distribution for each estimated future state and observation, whereas the term $\mathbf{H} [ \mathbb{E}_{Q(o_\tau, s_\tau)} Q(\pi | o_\tau, s_\tau)$ is defined as the entropy of the policy in the Gaussian mixture averaging all possible future observation and states. In our experiment, we minimize $- \mathbf{H} [ \mathbb{E}_{Q(o_\tau, s_\tau)} Q(\pi | o_\tau, s_\tau) $ similar to the reinforcement learning literature where the agent get some intrinsic reward when the policy entropy increases. The term $\mathbb{E}_{Q(s_\tau | \theta, \pi)} \mathbf{H} [\mathbb{E}_{Q(\theta,  \pi)} Q(o_\tau | s_\tau, \theta, \pi)] - \mathbb{E}_{Q(s_\tau | \theta, \pi) Q(\theta|\pi) } \mathbf{H} [ Q (o_\tau | s_\tau, \theta, \pi)]$ is ``the parameter exploration or active learning'' term~\cite{Schwartenbeck2019Curiosity, friston2017curiosity} in predicting the future observation, and the term $\mathbf{H} [ \mathbb{E}_{Q(\theta , \pi)} Q(s_\tau | \theta, \pi)] - \mathbb{E}_{Q(\theta|\pi)} \mathbf{H} [ Q (s_\tau | \theta, \pi)]$ is the ``parameter exploration or active learning'' term in predicting future states. Note that, since we do not specifically compute/rollout future observations due to impracticality, the parameter exploration over future observation is omitted.

Secondly, we decompose the second term of the expected free energy function as followed:

\begin{equation}
\begin{aligned}
    \mathbb{E}_{\Tilde{Q}} [ \ln Q(s_\tau | \theta, \pi) - \ln Q(s_\tau | o_\tau, \pi) ] & = \mathbb{E}_{Q(o_\tau, s_\tau, \theta | \pi)}[ \ln Q(s_\tau | \theta, \pi) - \ln Q(s_\tau | o_\tau, \pi)] \\
    & = \mathbb{E}_{Q(o_\tau | s_\tau, \theta, \pi) Q(\theta | \pi)} \mathbf{H} [ Q(s_\tau | \theta, \pi) ] - \mathbb{E}_{Q( \theta | o_\tau, s_\tau, \pi) Q(o_\tau | \pi)} \mathbf{H} [Q(s_\tau | o_\tau, \pi)].
\end{aligned}
\end{equation}

This term can be explained as the ``hidden state exploration or active inference''~\cite{Schwartenbeck2019Curiosity, friston2017curiosity}, which can be defined as the entropy of future prior state distribution rolled out from the world model minus the entropy of future state posterior estimated from future predicted observation. In our experiment, since we only roll-out from the state, and not observation, we can omit the term $- \mathbb{E}_{Q( \theta | o_\tau, s_\tau, \pi) Q(o_\tau | \pi)} \mathbf{H} [Q(s_\tau | o_\tau, \pi)]$. Additionally, instead of minimizing the entropy of future states $\mathbb{E}_{ Q(\theta | \pi)} \mathbf{H} [ Q(s_\tau | \theta, \pi) ]$, we maximize it to balance the omitted term and to give the agent more incentive in exploring un-visited states which has higher state entropy. This is also similar to providing agent with more motivation~\cite{Barto2013IntrinsicMotivation, Oudeyer2007IntrinsicMotivation, Deci1985IntrinsicMotivationBook} when there is a certain level of uncertainty estimated to be in future states. The ``hidden state exploration'' can then be approximately equal to $- \mathbb{E}_{ Q(\theta | \pi)} \mathbf{H} [ Q(s_\tau | \theta, \pi) ]$.

Lastly, for the third term, we are maximizing the probability of future predicted observation which match our prior preference at the same time step. Since closer states $s_\tau$ eventually leads to closer observation $o_\tau$, we can formulate this instrumental term as:

\begin{equation}
\begin{aligned}
    \ln P(o_\tau) & \approx r_\tau + \text{ sim} (s_\tau, \hat{s}_\tau)
\end{aligned}
\end{equation}
where $r_\tau$ is the predicted future reward observation, and $\text{sim} (s_\tau, s^*_\tau)$ is the similarity measure that we have defined in the manuscript. 

Overall, we have the full form of the expected free energy formula given the policy $\pi$:

\begin{equation}\label{eqn:G}
\begin{aligned}
    G_\tau(\pi) = &\ - \mathbf{H} [ \mathbb{E}_{Q(o_\tau, s_\tau)} Q(\pi | o_\tau, s_\tau)]\\
    & + \mathbf{H} [ \mathbb{E}_{Q(\theta , \pi)} Q(s_\tau | \theta, \pi)] - \mathbb{E}_{Q(\theta|\pi)} \mathbf{H} [ Q (s_\tau | \theta, \pi)]\\
    & - \mathbb{E}_{ Q(\theta | \pi)} \mathbf{H} [ Q(s_\tau | \theta, \pi) ] \\
    & - r_\tau - \text{ sim} (s_\tau, \hat{s}_\tau).
\end{aligned}
\end{equation}

\subsection{Discussion}

Being able to estimate a particular goal or preferred state at any particular time step is very useful for cognitive control agent. As our results demonstrate, the agent can then learn to adapt to take actions that lead from a specific state to its estimated goal(s). In contrast to the approaches taken in the imitation learning and behavior cloning literature, which train the agent's policy based on a fixed collected expert dataset, in our R-AIF framework, the expert signal (the preferred observation) is estimated dynamically through an adaptive prior preference model, closing the domain gap between the actual trajectories and the collected training imitation (preferred) trajectories.

Note that, with respect to our CRSPP sub-module, our model utilizes contrastive objective to adapt its parameters (i.e., it solely learns how to estimate the world model's encoded states while pushing itself away from undesired world model states/trajectories) and thus does \emph{not} require or learn any decoder. We only make use of the learned decoder in the generative world model to visualize the preferred observation from the estimated CRSPP's states. Experimentally, we remark that using an auxiliary decoder proved useful for clearly visualizing the preferred observation $\hat{o}_t$ given the produced preferred state $\hat{s}_t$ at each time step.

Theoretically, behavior cloning will be unable to achieve a great of success in multi-goal environments due to its reliance on a fixed, finite-size imitation sample pool. The expert trajectories in this fixed pool might further differ from the actual trajectories that the agent needs to take to solve the problem at hand, i.e., there is a distributional gap in observations, and therefore require different sets of actions to be taken. On the other hand, CRSPP learns to produce a dynamic goal state while jointly training the agent's core policy to reach task goal states. Therefore, as we empirically confirmed in simulation, R-AIF does not suffer from goal-mismatch problem in the training phase that other AIF schemes would.

\textbf{Broader Impacts.} In line with active inference process theory's focus on optimizing a policy that minimizes future expected free energy, R-AIF agents do so by taking actions that they are able to predict will lead to their preferences (in line AIF's instrumental signal), while also jointly taking actions that they are mostly sure about and working to reduce uncertainty by taking intelligent explorative actions of their niches (in line with AIF's epistemic signals). This is practical for different robotic tasks, particularly those with potential dangers in their operation/functioning (i.e., when human safety must be considered). For example, a self-driving car agent can take the actions that it is sure to be safe rather than focusing exploring wildly (as random exploration policies encourage, potentially causing traffic accidents. Furthermore, the R-AIF framework could prove useful to the imitation learning research community, as its ability to optimize a policy that achieves dynamic goals as produced from an adaptive prior preference model was found to be quite useful for the more complex POMDP tasks we sought to solve in this work. This carries with it possible positive implications for practical application in downstream tasks such as autonomous driving~\cite{Nozari2022AIFAutonomousDriving} and complex robotic control and navigation~\cite{Krayani2022AIFUAV, Oliver2022aifhumanrobot}.


\end{document}